\documentclass[default,iicol]{sn-jnl}

\usepackage{graphicx}%
\usepackage{multirow}%
\usepackage{amsmath,amssymb,amsfonts}%
\usepackage{amsthm}%
\usepackage{mathrsfs}%
\usepackage[title]{appendix}%
\usepackage{xcolor}%
\usepackage{textcomp}%
\usepackage{manyfoot}%
\usepackage{booktabs}%
\usepackage{algorithm}%
\usepackage{algorithmicx}%
\usepackage{algpseudocode}%
\usepackage{listings}%

\usepackage{placeins}
\usepackage{array}
\usepackage{stfloats}
\usepackage{url}
\usepackage{verbatim}
\usepackage{colortbl}  
\usepackage{makecell}
\usepackage{tabularx}
\usepackage{float}

\theoremstyle{thmstyleone}%
\theoremstyle{thmstyletwo}%

\theoremstyle{thmstylethree}%
\definecolor{c1}{HTML}{FFEBEB}
\definecolor{c2}{HTML}{FFF2E0}
\definecolor{c3}{HTML}{FFFFD6}
\definecolor{c4}{HTML}{FBD9FC}
\definecolor{c5}{HTML}{D8D3FE}
\definecolor{c6}{HTML}{B0D6FF}
\definecolor{c7}{HTML}{91B2F2}
\definecolor{c8}{HTML}{FFEBEB}
\definecolor{c9}{HTML}{FFF2E0}
\definecolor{c10}{HTML}{FFFFD6}

\begin{document}

\title[A Survey on Personalized Content Synthesis with Diffusion Models]{A Survey on Personalized Content Synthesis with Diffusion Models}


\author[1,2]{\fnm{Xulu} \sur{Zhang}}

\author*[1]{\fnm{Xiaoyong} \sur{Wei}}\email{x1wei@polyu.edu.hk}

\author[1]{\fnm{Wentao} \sur{Hu}}

\author[2,3]{\fnm{Jinlin} \sur{Wu}}

\author[1]{\fnm{Jiaxin} \sur{Wu}}

\author[1]{\fnm{Wengyu} \sur{Zhang}}

\author[2,3,4]{\fnm{Zhaoxiang} \sur{Zhang}}

\author*[2,3,4]{\fnm{Zhen} \sur{Lei}}\email{zhen.lei@ia.ac.cn}

\author[1]{\fnm{Qing} \sur{Li}}


\affil[1]{\orgdiv{Department of Computing}, \orgname{The Hong Kong Polytechnic University}, \orgaddress{\city{Hong Kong} \postcode{999077},  \country{China}}}

\affil[2]{\orgdiv{Center for Artificial Intelligence and Robotics}, \orgname{Hong Kong Institute of Science \& Innovation, Chinese Academy of Sciences}, \orgaddress{\city{Hong Kong} \postcode{999077},  \country{China}}}


\affil[3]{\orgdiv{State Key Laboratory of Multimodal Artificial Intelligence Systems}, \orgname{Chinese Academy of Sciences Institute of Automation}, \orgaddress{\city{Beijing} \postcode{100190},  \country{China}}}

\affil[4]{\orgdiv{School of Artificial Intelligence}, \orgname{University of Chinese Academy of Sciences}, \orgaddress{\city{Beijing} \postcode{100049},  \country{China}}}


\vspace{-2in}
\abstract{Recent advancements in diffusion models have significantly impacted content creation, leading to the emergence of Personalized Content Synthesis (PCS).  
By utilizing a small set of user-provided examples featuring the same subject, PCS aims to tailor this subject to specific user-defined prompts.
%
Over the past two years, more than 150 methods have been introduced in this area.
However, existing surveys primarily focus on text-to-image generation, with few providing up-to-date summaries on PCS. 
This paper provides a comprehensive survey of PCS, introducing the general frameworks of PCS research, which can be categorized into test-time fine-tuning (TTF) and pre-trained adaptation (PTA) approaches.
%
%
We analyze the strengths, limitations, and key techniques of these methodologies.
%
Additionally, we explore specialized tasks within the field, such as object, face, and style personalization, while highlighting their unique challenges and innovations.
Despite the promising progress, we also discuss ongoing challenges, including overfitting and the trade-off between subject fidelity and text alignment. Through this detailed overview and analysis, we propose future directions to further the development of PCS.
}




\maketitle

\footnotetext{Accepted by Machine Intelligence Research, DOI: 10.1007/s11633-025-1563-3}
\section{Introduction}\label{sec1}

\begin{figure}[t]
    \centering
    \includegraphics[width=1.0\linewidth]{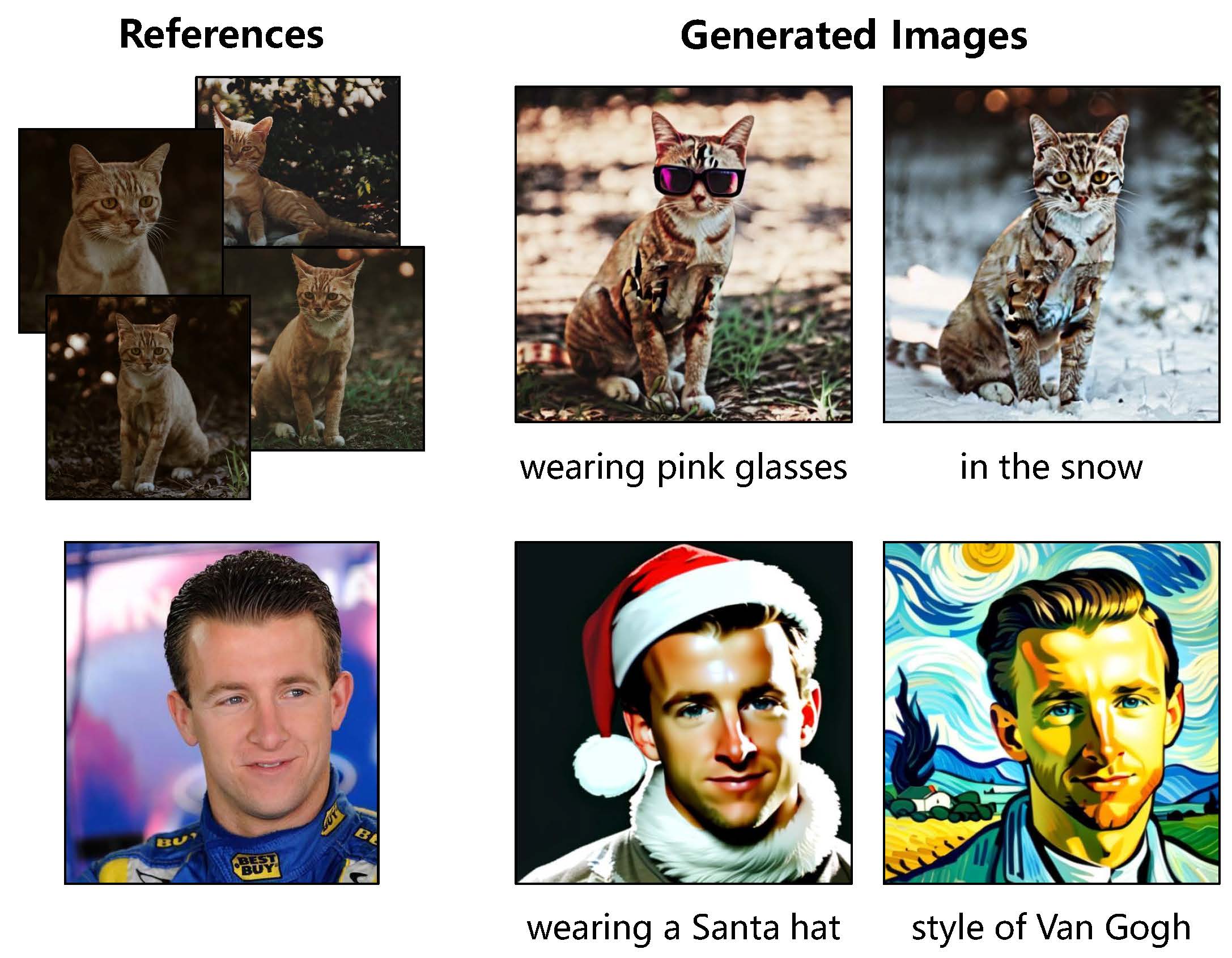}
    \caption{Given a few reference images of a subject (e.g., a cat \cite{2ruiz2023dreambooth} or face \cite{54xiao2023fastcomposer}), PCS aims to generate new renditions of the subject that align with user-defined textual prompts. The task requires preserving the subject’s identity while adapting to diverse contexts. The examples are generated by this survey using DreamBooth \cite{2ruiz2023dreambooth} and InstantID \cite{37wang2024instantid}.}
    \label{fig:example}
\end{figure}

Recently, generative models have shown remarkable progress in the field of natural language processing and computer vision.
Notable examples, such as ChatGPT \cite{wu2023brief} and text-to-image diffusion models \cite{croitoru2023diffusion}, have showcased impressive capabilities in content creation.
However, these advanced models often struggle to fulfill specific requirements, such as answering domain-specific queries or accurately depicting user's portraits. 
This limitation highlights the importance of Personalized Content Synthesis (PCS), which enables users to customize models for their unique tasks and requirements. 
As a critical area of research, PCS is increasingly recognized as essential for achieving Artificial General Intelligence (AGI), evidenced by the growing number of companies releasing products that support personalized content creation, like utilizing Reinforcement Learning to fine-tune language models \cite{uc2023survey}.
In this paper, we focus on PCS within the context of diffusion models in computer vision. The objective of PCS is to learn the subject of interest (SoI) from a small set of user-uploaded samples and generate images that align with user-defined contexts.
For instance, as illustrated in Fig.~\ref{fig:example}, PCS can customize a unique cat into various scenarios, such as ``wearing pink sunglasses'' and ``in the snow''.

\begin{figure*}[t]
    \centering
    \includegraphics[width=0.95\textwidth]{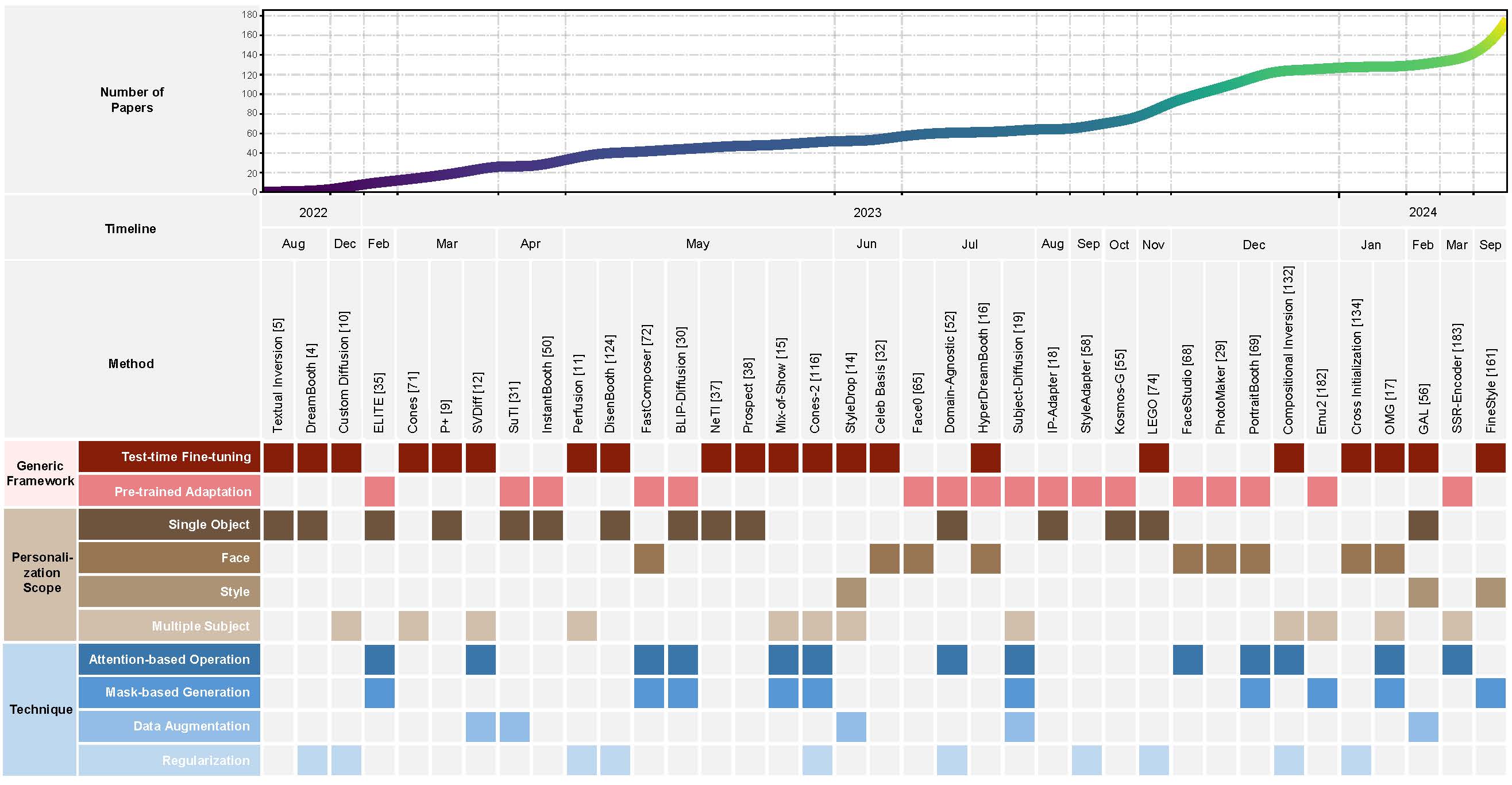}
    \caption{A chronological overview of classical PCS methods as surveyed, illustrating the evolution of techniques through months. The number of related works has rapidly increased over the past two years. We divide PCS methods with 3 different criteria: training strategy, personalization scope, and technique.}
    \label{fig:timestone}
\end{figure*}

The emergence of diffusion models has significantly facilitated text-guided content generation, leading to a rapid expansion of PCS. As depicted in Fig.~\ref{fig:timestone}, the number of research papers on PCS has surged over time. Following the release of key innovations such as DreamBooth \cite{2ruiz2023dreambooth} and Textual Inversion \cite{1gal2022image} in August 2022, over 150 methods have been proposed in a remarkably short period.
These methods can be classified using several criteria. First, in terms of \textbf{training strategy}, we differentiate between test-time fine-tuning (TTF) and pre-trained adaptation (PTA) approaches. TTF methods fine-tune generative models for each personalization request during inference phase, while PTA methods strive to pre-train a unified model capable of generating a wide range of subjects of interest (SoI).
Second, we categorize \textbf{the techniques employed in PCS} into four primary areas: attention-based operations, mask-guided generation, data augmentation, and regularization. As shown in Fig.~\ref{fig:timestone}, we summarize various influential PCS methods based on these criteria.
Third, \textbf{the scope of personalization} has expanded considerably. The SoI now includes not only general objects but also extends to human faces, artistic styles, actions, and other intricate semantic elements. This growing interest encompasses the creation of complex compositions and the combination of several conditions. Additionally, the research landscape has broadened from static images to other modalities such as video and 3D representations.
%

\begin{figure}[ht]
    \centering
    \includegraphics[width=0.98\linewidth]{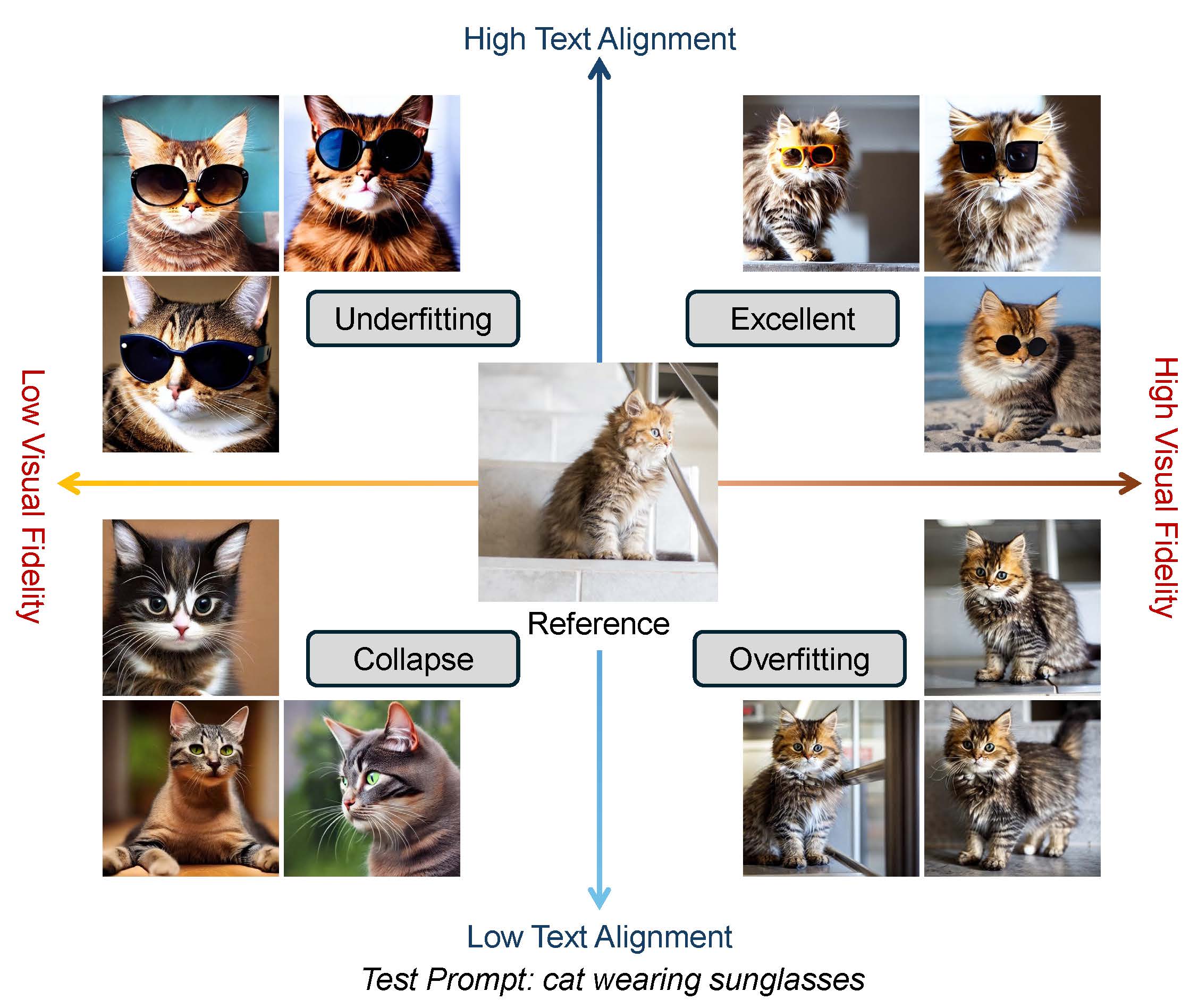}
    \caption{The trade-off between text alignment and visual fidelity in personalized image synthesis, illustrated through DreamBooth-generated \cite{2ruiz2023dreambooth} examples of a customized cat wearing sunglasses. Overfitting occurs when the model focuses solely on reconstructing the cat, disregarding the sunglasses context. Underfitting, on the other hand, reflects the model's attempt to satisfy the text prompt but fails to accurately represent the personalized cat. Collapse signifies a failure to meet both criteria.}
    \label{fig:challenge}
\end{figure}

While current methods in PCS have shown impressive performance, several challenges remain unresolved. A primary concern is the issue of overfitting, which often arises from the limited number of reference images available. This limitation can lead to a neglect of the textual context in the generated outputs, as shown in Fig.~\ref{fig:challenge}. 
%
Another related challenge is the trade-off between image alignment and text fidelity, as illustrated in Fig.~\ref{fig:challenge}. When a model successfully reconstructs the fine-grained details of the subject of interest (SoI), it often sacrifices controllability. Conversely, enhancing the model's editability can result in underfitting with a loss of fidelity to the original SoI.
%
Additionally, there are other challenges including the absence of robust evaluation metrics, a lack of standardized test datasets, and the need for faster processing times. 
%
This paper discusses these challenges in detail and establishes a benchmark for evaluating classical methods. By addressing these obstacles, we aim to advance the field of personalized content synthesis and enhance its practical applications.

This survey aims to provide a comprehensive overview of the existing methods, frameworks, and challenges of PCS to help readers understand current trends in this evolving landscape and promote further improvements in this area.
The structure of this survey is organized into several key sections: 
In Section \ref{sec:fundamentals}, We begin with a brief introduction to diffusion models for better understanding of PCS.
In Section \ref{sec:framework}, we introduce two primary frameworks for PCS: test-time fine-tuning and pre-trained adaptation.
In Section \ref{sec:techniques}, we categorize common techniques into four main parts, attention-based methods, mask-based generation, data augmentation, and regularization.
In Section \ref{task}, we summarize various PCS methods based on different image personalization tasks, including personalization on object, face, style, and etc.
In Section \ref{sec:extension}, we demonstrate PCS's growing impact in video, 3D, and other areas beyond traditional image generation tasks.
In Section \ref{evaluation}, we review the current evaluation dataset and metrics, and introduce a benchmark on the existing methods to promote further research.
In Section \ref{challenge}, we discuss current challenges faced in PCS and propose potential future directions in this area.
Additionally, we present a comprehensive summary of all relerated papers in Table \ref{tb:summary1}, Table \ref{tb:summary2}, and Table \ref{tb:summary3}.

\section{Fundamentals}  
\label{sec:fundamentals}  
Modern diffusion modeling has evolved into a sophisticated framework that unifies discrete and continuous generative paradigms through stochastic differential equations (SDEs) \cite{song2020score}, ordinary differential equations (ODEs) \cite{song2020score,karras2022elucidating}. This section introduces the core mathematical foundations, emphasizing innovations in the diffusion process and conditional mechanisms. By covering these theoretical developments, this section lays the groundwork for the personalization tasks explored in subsequent sections.

\subsection{Denoising Diffusion Probabilistic Models (DDPMs)}  
The architecture of diffusion models comprises two complementary processes: a \textit{forward diffusion} that systematically perturbs data distributions, and a \textit{learned reverse process} that reconstructs signals through iterative refinement. Originating from discrete-time Markov chains in Denoising Diffusion Probabilistic Model (DDPM) \cite{ho2020denoising}, the forward process applies Gaussian noise corruption over \( T \) steps:  

\begin{align}
\mathbf{x}_t = \sqrt{\bar{\alpha}_t}\mathbf{x}_0 + \sqrt{1-\bar{\alpha}_t}\mathbf{\epsilon}, \quad \bar{\alpha}_t = \prod_{s=1}^t \alpha_s  
\end{align}

where \(\mathbf{x}_t\) denotes the noised data at diffusion step \(t\), $\mathbf{\epsilon} \sim \mathcal{N}(0,\mathbf{I})$ is the standard Gaussian noise vector that corrupts the original data $\mathbf{x}_0$, and \(\alpha_s\) governs the noise scheduling policy.

The generative capability resides in learning to invert this degradation through parameterized reverse transitions. For DDPMs, this inversion is achieved via Bayesian reconstruction:

\begin{multline}
p_{\theta}\left(\mathbf{x}_{t-1}\mid\mathbf{x}_t\right) = \\ 
\mathcal{N}\left(\mathbf{x}_{t-1}; \frac{1}{\sqrt{\alpha_t}}\left(\mathbf{x}_t - \frac{\beta_t}{\sqrt{1-\bar{\alpha}_t}}\epsilon_\theta(\mathbf{x}_t,t)\right),\beta_tI\right)
\end{multline}

Here, \(p_{\theta}\) represents the parameterized reverse transition probability, \(\epsilon_\theta\) is the neural network predicting noise components, and \(\beta_t\) is a hyperparameter before model training. And the neural network \(\epsilon_\theta\) learns to predict the applied noise component \(\epsilon_\theta\) through:

\begin{align}
\mathcal{L} = \mathbb{E}_{\mathbf{x}_0,t,\mathbf{\epsilon}}\left[\|\mathbf{\epsilon} - \mathbf{\epsilon}_\theta(\mathbf{x}_t,t)\|_{2}^2\right]  
\end{align} 

This objective enables end-to-end training of the denoising trajectory. 

Once training is finished, the model can apply the learned reverse diffusion transitions to any arbitrary noise input, progressively denoising it into a coherent data sample. This capability allows DDPMs to function as general-purpose generative engines, producing high-quality outputs from random noise vectors.

\subsection{Stochastic Differential Equations (SDEs)}
Although DDPMs establish basic denoising dynamics, this discrete formulation imposes constraints on flexible noise scheduling and computational efficiency.  

To overcome these limitations, the framework was generalized through continuous-time SDE \cite{song2020score}, which provide a unified perspective for modeling diffusion dynamics: 

\begin{align}
d\mathbf{x} = \mathbf{f}(\mathbf{x},t)dt + g(t)d\mathbf{w}  
\label{eq:SDEs}
\end{align}

where \(\mathbf{f}(\mathbf{x},t)\) encodes deterministic drift components, \(g(t)\) modulates stochastic diffusion via Wiener process \(\mathbf{w}\). This continuous perspective subsumes DDPMs as special cases while enabling adaptive noise scheduling strategies such as variance-preserving (VP) and variance-exploding (VE) schedules through strategic choices of \(\mathbf{f}\) and \(g\) \cite{karras2022elucidating}.  

For the reverse process, a crucial result \cite{anderson1982reverse} shows that the reverse diffusion process also follows an SDE:

\begin{align}
d\mathbf{x} = \left[\mathbf{f}(\mathbf{x},t) - g(t)^2\nabla_\mathbf{x}\log p_t(\mathbf{x})\right]dt + g(t)d\mathbf{\bar{w}}  
\end{align}

where \(\bar{w}\) denotes reverse-time Wiener increments, and \(\nabla_\mathbf{x}\log p_t(\mathbf{x})\) is a score function. This formulation enables generation by integrating backward from \(\mathbf{x}_T \sim \mathcal{N}(0,I)\) to \(\mathbf{x}_0\), conditioned on learned score estimates.

\subsection{Ordinary Differential Equations (ODEs)}  
While SDEs provide a comprehensive framework for modeling diffusion dynamics, their inherent randomness introduces critical challenges in practical deployment. The Wiener process \(\mathbf{w}\) in SDE-based sampling generates pathwise variability, requiring extensive Monte Carlo averaging to converge to stable solutions. Furthermore, stochastic trajectories exhibit erratic curvature profiles, forcing fixed-step solvers to adopt conservative discretization schemes that often necessitate 1,000+ steps for high-fidelity generation.  

These limitations motivate the derivation of deterministic sampling trajectories through probability flow ordinary differential equations (ODEs) \cite{song2020score,karras2022elucidating}, obtained by eliminating the stochastic term from the SDE formulation:  

\begin{align}
d\mathbf{x} = \left[\mathbf{f}(\mathbf{x},t) - \frac{1}{2}g(t)^2\nabla_\mathbf{x}\log p_t(\mathbf{x})\right]dt
\end{align} 

The ODE’s deterministic nature arises from two synergistic components: the original drift term \(\mathbf{f}(\mathbf{x},t)\) and a correction term involving the score function \(\nabla_\mathbf{x}\log p_t(\mathbf{x})\). This score-guided adjustment preserves the marginal data distribution \(p_t(\mathbf{x})\) while eliminating pathwise stochasticity. Crucially, the resulting trajectories exhibit intrinsic geometric regularity, allowing the complex denoising process to operate in a simplified mathematical space. This enabling 5-10x faster sampling than SDEs through adaptive ODE solvers like DPM-Solver \cite{lu2022dpm} and DPM-Solver++ \cite{lu2022dpmplus}.  
Recent work \cite{chen2024trajectory} further identifies strong geometric regularity in these ODE-based sampling trajectories, demonstrating that they inherently follow a linear-nonlinear-linear structure regardless of generated content. This trajectory regularity enables dynamic programming-based time scheduling optimization with negligible computational overhead.

\subsection{Conditional Generation Mechanisms}
Recently, conditional synthesis has emerged as the critical capability bridging theoretical diffusion frameworks with real-world applications. It enables precise alignment of outputs with multimodal guidance signals (text prompts, subject embeddings, anatomical masks), fulfilling domain-specific requirements for reliability and reproducibility.

Building upon the unconditional framework, conditional synthesis is formalized through extended score matching:  

\begin{align}
    \mathcal{L} = \mathbb{E}_{\mathbf{x}_0,t,\mathbf{\epsilon}}\left[\|\mathbf{\epsilon} - \mathbf{\epsilon}_\theta(\mathbf{x}_t,t,c)\|_{2}^2\right]  
    \label{eq:condition_generation}
\end{align}

where the conditioning signal \( c \)  can be integrated via multiple synergistic mechanisms, such as cross-modal attention \cite{rombach2022high} and spatial modulation \cite{zhang2023adding}.

This conditional paradigm directly enables the applications of PCS. Leading text-to-image systems, such as Stable Diffusion (SD) \cite{rombach2022high} and DALLE \cite{ramesh2022hierarchical}, are widely adopted to empower users in controlling customized content through text instructions.

\section{Generic Framework}
\label{sec:framework}
In this survey, we broadly categorize PCS frameworks into two paradigms: TTF and PTA approaches.
These two frameworks fundamentally differ in their adaptation mechanisms.
TTF methods dynamically adjust model parameters for each new subject during inference, prioritizing visual fidelity at the cost of computational overhead.
Conversely, PTA frameworks employ reference-aware architectures trained on large datasets to enable single-pass personalization without parameter updates during inference.
We introduce two generic frameworks in the following sections.

\subsection{Test-time Fine-tuning (TTF) Framework}

\begin{figure}[h]
    \centering
    \includegraphics[width=1.0\linewidth]{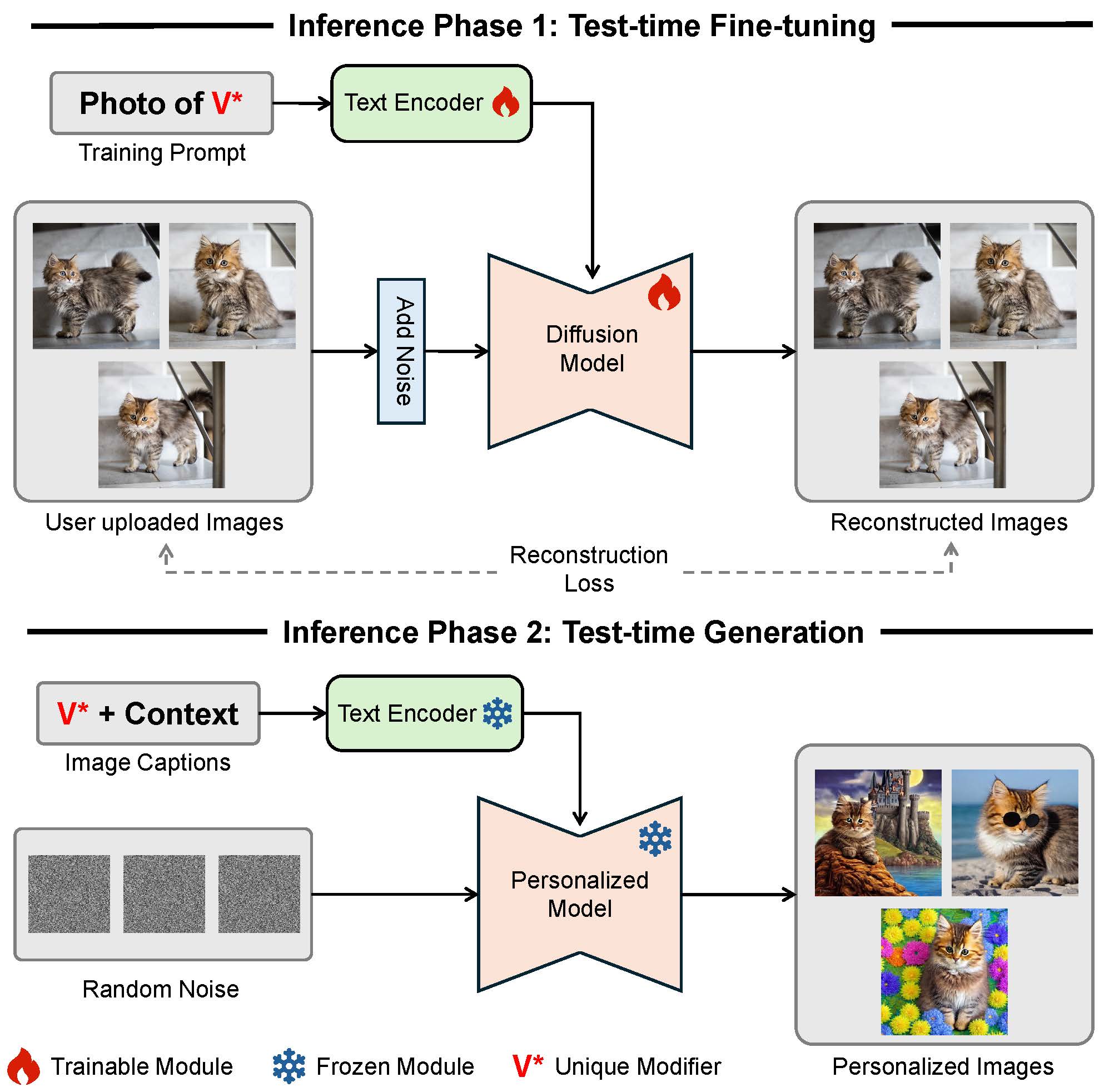}
    \caption{Illustration of the TTF framework for the test-time fine-tuning process and generation phase. During the inference phase, the model fine-tunes its parameters by reconstructing the reference images for each SoI group. The unique modifier \textit{V*} is employed to represent the SoI and used to formulate new inference prompts for generating personalized images.}
    \label{fig:optimization}
\end{figure}



TTF methods represent the foundational approach for personalized synthesis, where models adapt to new subjects through instance-specific optimization during inference.
As illustrated in Fig.~\ref{fig:optimization}, this framework operates through two core principles:
1) \textbf{Test-time adaptation} fine-tunes model parameters to learn the key visual element of the SoI.
2) \textbf{Semantic-aware modifier system} represents the SoI at the token level to bridge the gap between visual adaptation and textual control.
In the following Section \ref{sec:subsubTTF} and Section \ref{sec:unique_modifier}, we will detail these principles respectively.

\subsubsection{Test-time Adaption}
\label{sec:subsubTTF}
\textbf{Test-time fine-tuning}. For each reference set \(\mathbf{X}_{SoI}\) of SoI, the optimization process adjusts a subset of parameters $\mathbf{\theta^{\prime}}$ to reconstruct the SoI conditioned on the reference prompt.
The fine-tuning objective is defined by a reconstruction loss:

\begin{align}
\mathcal{L} = \mathbb{E}_{\mathbf{x}_0\in\mathbf{X}_{SoI},t,\mathbf{\epsilon}}\left[\|\mathbf{\epsilon} - \mathbf{\epsilon}_{\theta^{\prime}}(\mathbf{x}_t,t,c)\|_{2}^2\right] 
      \label{eq:optimization_recons_loss_text}
\end{align}

where \(c\) represents the condition signal, typically the reference image caption.
Compared to the large-scale pre-training described in Equation \ref{eq:condition_generation}, key differences lie in the training data and learnable parameters.
The training samples are typically restricted to the SoI references, sometimes supplemented with a regularization dataset to mitigate overfitting \cite{2ruiz2023dreambooth}.
%
%
For the selection of learnable parameters $\mathbf{\theta^{\prime}}$, the commonly adopted options include token embeddings \cite{1gal2022image,23voynov2023p+}, the entire diffusion model \cite{2ruiz2023dreambooth,151zhang2024generative}, specific subsets of parameters \cite{6kumari2023multi,7tewel2023key,8han2023svdiff}, or introducing new parameters such as adapters \cite{72xiang2023closer,5sohn2024styledrop}, and LoRA \cite{46gu2024mix,58ruiz2023hyperdreambooth,115kong2024omg}, which will be discussed in Section \ref{sec:parameter}.

\textbf{Test-time generation}. The test-time generation is proceeded once the model has been fine-tuned with the optimized parameters $\mathbf{\theta^{\prime}}$. By composing novel input prompts that incorporate the SoI's unique identifier (introduced in Section \ref{sec:unique_modifier}), the adapted model can synthesize diverse images while preserving the subject's distinctive characteristics. This approach enables flexible control over the generated content through natural language descriptions.

\subsubsection{Unique Modifier}
\label{sec:unique_modifier}
A unique modifier is a textual token or short phrase that uniquely represents a SoI, enabling text-based representation of the SoI for flexible prompt composition. 
As illustrated in Fig.~\ref{fig:optimization}, this modifier serves as a textual description for the SoI, allowing combination with other descriptions (e.g., ``\textit{V*} on the beach") during inference.
Normally, the construction of the unique modifier can be divided into three categories:

\textbf{Plain text} \cite{2ruiz2023dreambooth,6kumari2023multi,7tewel2023key,5sohn2024styledrop,151zhang2024generative}. This approach utilizes an explicit text description to represent the SoI. For example, words such as \textit{cat} could directly represent the user's cat in the references. This setting typically requires fine-tuning the parameters of the diffusion model components, such as UNet \cite{ronneberger2015unet} or Transformer blocks \cite{vaswani2017attention}, to enable the model to associate the SoI’s visual features with the plain‐text token. The plain text offers user-friendly prompt construction and injects subject prior information to ease tuning difficulty. However, this technique may over-specialize common terms, limiting their broader applicability as the model learns to associate general words with specific SoI characteristics.

\textbf{Rare token} \cite{2ruiz2023dreambooth}. This technique employs infrequently used tokens to minimize their impact on commonly used vocabulary. Similar to the plain text approach, the embeddings of rare tokens remain unchanged during fine-tuning. However, these rare tokens often fail to provide useful subject prior information and still exhibit weak interference with unrelated vocabulary, potentially leading to ambiguity between the original meaning and the intended SoI reference.

\textbf{Learnable token embedding} \cite{1gal2022image,23voynov2023p+,61alaluf2023neural,30zhang2023prospect,48wang2023hifi,66pang2023cross}. This method adds a new token and its corresponding embedding vector to the dictionary of the tokenizer. An intuitive example is that this method creates a new word that does not exist in the dictionary. This inserted token has adjustable weights during fine-tuning, while the embeddings of other tokens in the pre-defined dictionary will remain unchanged. This approach requires only a few kilobytes of additional parameters and maintains the base model's capabilities for non-customized generation. Similar to the rare token approach, it is not very user-friendly in practice, since users must learn and remember an unfamiliar token to reference their subject of interest.

\subsubsection{Training Parameter Selection}
\label{sec:parameter}
The selection of trainable parameters represents a critical design consideration in PCS, directly impacting many key performance metrics: subject fidelity, training efficiency, and model storage requirements. 
Current tunable parameters can be broadly categorized into the following four types. 

\textbf{Token embedding}. As introduced in Section \ref{sec:unique_modifier}, token embedding optimization \cite{1gal2022image} introduces learnable tokens as the unique modifier to represent SoI through noise-to-image reconstruction. While achieving remarkable parameter efficiency, this approach faces challenges in detail preservation and prolonged training time (typically larger than 20 minutes) due to the inherent compression of complex features into low-dimensional embeddings. Subsequent works \cite{23voynov2023p+,61alaluf2023neural,30zhang2023prospect,48wang2023hifi,66pang2023cross} aim to address these limitations through different strategies, which are summarized in Section \ref{sec:object}.

\textbf{Existing model parameters}. This paradigm directly optimizes pre-trained model components, such as the text encoder, U-Net blocks, and transformer layers \cite{2ruiz2023dreambooth,151zhang2024generative,6kumari2023multi,7tewel2023key,8han2023svdiff}. Benefitting from the advanced representation capacity of these modules, the fine-tuning phase can achieve faster convergence (5-10 minutes) and superior visual fidelity compared to token-only methods, though at the cost of significant storage overhead. In addition, these modules inherently support attention mechanisms to facilitate feature enhancement operations.

\textbf{Parameter-efficient extensions}. Recent advanced methods have introduced parameter-efficient techniques into PCS, such as LoRA \cite{46gu2024mix,58ruiz2023hyperdreambooth,115kong2024omg,51lu2024object,95achlioptas2023stellar,76yeh2023navigating} and adapter modules \cite{72xiang2023closer,5sohn2024styledrop}, which inject small, trainable components into the base model. These methods achieve comparable performance to full parameter fine-tuning while dramatically reducing storage requirements.

\textbf{Combined strategy}. Since the aforementioned strategies are not conflicting, some methods allocate different learning rates and training phases to each component type to achieve optimal balance between fidelity and efficiency.
For instance, the fine-tuned token embedding can be regarded as an effective initialization for the subsequent model weight fine-tuning \cite{90hyung2023magicapture}. 
Also, these two parts can be simultaneously optimized with different learning rates \cite{74avrahami2023break,86hua2023dreamtuner}.

\subsubsection{Prompt Engineering}
The construction of training prompts for samples typically starts with adding prefix words before the modifier token. A simplest example is ``Photo of \textit{V*}".
However, DreamBooth \cite{2ruiz2023dreambooth} noted that such a simple description causes a long training time and unsatisfactory performance.
To address this, they incorporate the unique modifier with a class noun to describe the SoI in the references (e.g., ``Photo of \textit{V*} cat").
Also, the training caption for each training reference can be more precious for better disentanglement of SoI and irrelevant concepts \cite{57he2023data}, such as ``Photo of \textit{V*} cat on the chair".
This follows the tending that high-quality captions in the training set could assist in further improvement of accurate text control \cite{BetkerImprovingIG}.

The PTA framework has emerged as a breakthrough approach for PCS, aiming to eliminate the computational burden of per-request fine-tuning while maintaining high-quality, subject-specific generation capabilities. 
To achieve this goal, this approach combines large-scale pretraining with reference-aware architectures to enable single-pass personalization, as shown in Fig.~\ref{fig:learning}.
Based on this architecture, three critical design factors are considered to ensure practical viability:
1) \textbf{Preservation of semantic-critical features} to guarantee visual consistency with reference inputs,
2) \textbf{Effective fusion} that combines reference features with text guidance to achieve desired generation,
3) \textbf{Optimization of training dataset} scale to achieve robust generalization without overfitting.
In the following sub-section, we first present overall architecture introduction in Section \ref{sec:subsubPTA} and then delve into three detailed factors.

\subsubsection{Pre-trianed Adaption}
\label{sec:subsubPTA}

\subsection{Pre-trained Adaptation (PTA) Framework}
\begin{figure}[ht]
    \centering
    \includegraphics[width=1.0\linewidth]{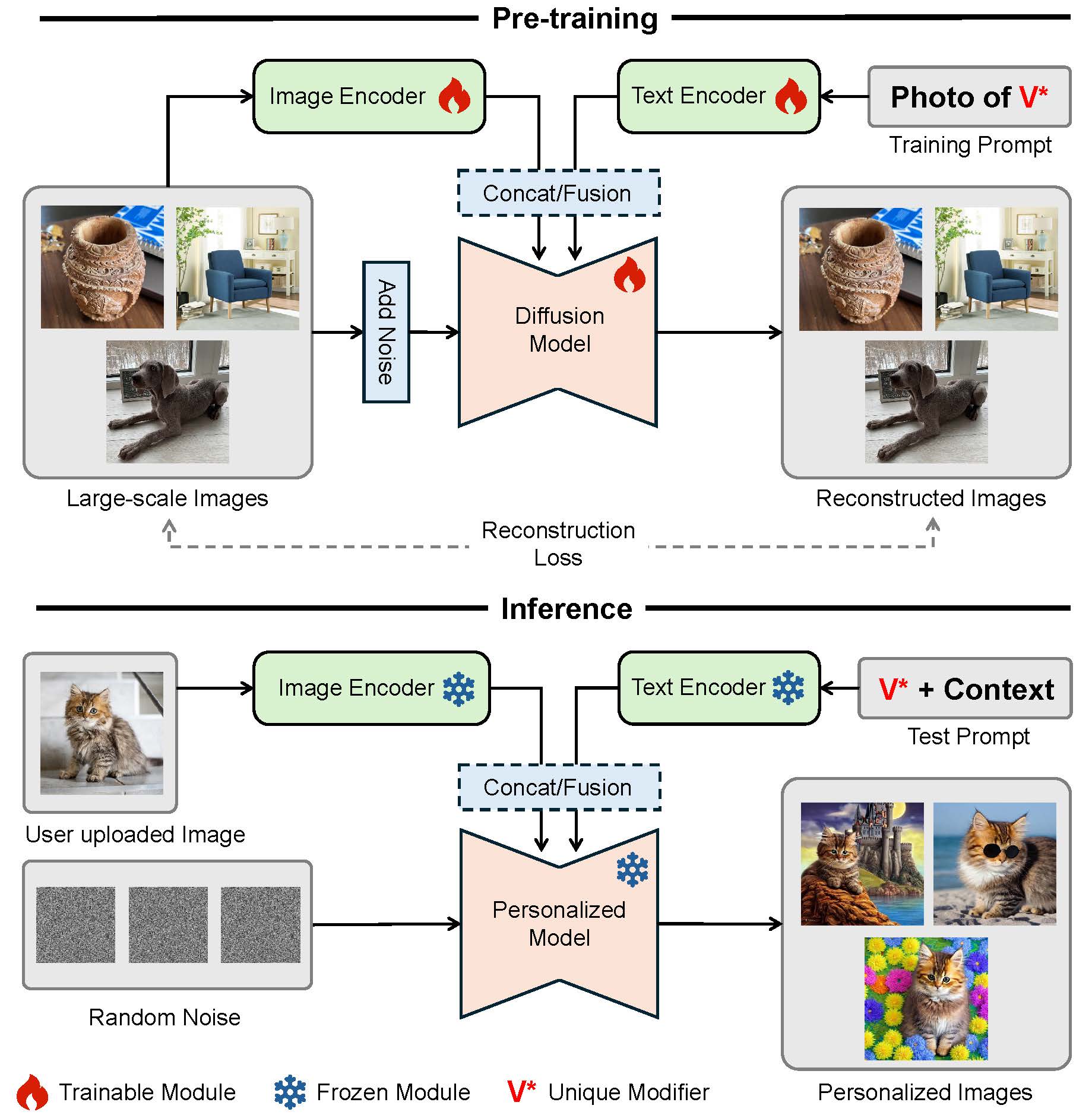}
    \caption{Illustration of the PTA method for personalized image synthesis. This framework utilizes a large-scale dataset to train a unified model that can process diverse personalization requests. The diffusion model is adapted to process hybrid inputs derived from both visual and textual features. Additionally, the concatenation of image and text features can be implemented in various ways, such as placeholder-based and reference-conditioned.}
    \label{fig:learning}
\end{figure}

\textbf{Pre-training}.
During the pre-training phase, the PTA framework aims to establish direct mappings between reference characteristics (e.g., facial features, object textures) and synthesized outputs.
To achieve this, the reference inputs are processed through dedicated feature extractors and fused with text prompts to serve as conditions to guide the generation, as shown in Fig.~\ref{fig:learning}.
A reconstruction loss is optimized that enforces the alignment between generated images and the large-scale dataset $X_{data}$:

\begin{align}
\mathcal{L} = \mathbb{E}_{\mathbf{x}_0\in\mathbf{X}_{data},t,\mathbf{\epsilon}}\left[\|\mathbf{\epsilon} - \mathbf{\epsilon}_{\theta^{\prime}}(\mathbf{x}_t,t,c^{\prime})\|_{2}^2\right] , \quad c^{\prime}=\mathcal{F}(c,\mathbf{x}_0)
      \label{eq:learning_recons_loss_text}
\end{align}

where $\mathcal{F}$ represents the fusion operation combining text condition \(c\) and reference image \(\mathbf{x}_0\). The tunable parameters \(\theta^{\prime}\) include visual encoder weights, text encoder components, diffusion modules, and injected adapter modules.

\textbf{Inference}. During inference, the PTA framework processes a reference image through its visual encoder to extract discriminative features, which are then fused with text embeddings using the pretrained conditioning module. This fused representation guides the diffusion model to generate personalized outputs. This approach effectively eliminates test-time optimization to ensure fast generation.

\subsubsection{Subject Feature Extraction}
%
%
%
Extracting representative features of the SoI is crucial in the creation of personalized content. 
A common approach is to employ an encoder, leveraging pre-trained models such as CLIP \cite{radford2021learning} and BLIP \cite{li2022blip}. 
%
While these models excel at capturing global features, they often include irrelevant information that can detract from the fidelity, potentially compromising the quality of the personalized output, such as including the same background in the generation.
To mitigate this issue, some studies incorporate additional prior knowledge to guide the learning process so as to focus on the targeted SoI. 
For instance, the SoI-specific mask \cite{21zhang2023personalize,48wang2023hifi,74avrahami2023break,81safaee2023clic,51lu2024object,95achlioptas2023stellar} contributes to the effective exclusion of the influence of the background. 
Moreover, using facial landmarks \cite{37wang2024instantid} in the context of human face customization helps improve identity preservation. 
We will discuss a more detailed technical summary of mask-assistant generation in Section \ref{sec:mask}.

Handling multiple input references presents another challenge but is essential for real-world deployment.
This necessitates an ensemble of features from the multiple reference images to augment the framework's adaptability.
Yet, the majority of current PTA systems are limited to supporting one reference input.
Some research works \cite{37wang2024instantid,28li2023photomaker} propose to average or stack features extracted from multiple references to form a composite SoI representation.

\subsubsection{Subject Feature Fusion}
%
Personalized content synthesis systems typically process two input modalities: reference images and textual descriptions, as illustrated in Fig.~\ref{fig:learning}. Effective fusion of these heterogeneous features represents a critical technical challenge in PTA frameworks. Current methodologies can be categorized into four primary approaches:

\textbf{Concatenation-based Fusion} \cite{19wei2023elite,28li2023photomaker,54xiao2023fastcomposer,89chen2023dreamidentity,85cai2023decoupled,96peng2023portraitbooth}. This method makes the unique modifier a placeholder token to encapsulate visual subject characteristics. The placeholder token embedding, initialized with image features from visual encoders, is concatenated with text embeddings from the language model. This combined representation subsequently guides generation through standard cross-attention layers in the diffusion process, enabling basic subject-text alignment while maintaining architectural simplicity.

\textbf{Cross-attention Fusion} \cite{40ye2023ip,11chen2024subject,37wang2024instantid,91li2023stylegan,212liu2024stylecrafter,221zhang2024ssr,220personalizationms}. This paradigm extends the U-Net architecture with specialized attention mechanisms that jointly process visual and textual conditions.
For example, IP-Adapter \cite{40ye2023ip} introduces decoupled cross-attention layers that maintain separate query projections for image and text features.
In this case, the unique modifier directly displays the plain text.

\textbf{Multimodal Encoder Fusion} \cite{10li2024blip,84zhou2023customization,song2024moma,77pan2023kosmos,71ma2023unified}. This approach leverages powerful multimodal encoder architectures (e.g., BLIP-2 \cite{li2023blip}) to jointly embed visual and textual subject descriptors. BLIP-Diffusion \cite{10li2024blip} exemplifies this strategy by learning a compact subject prompt embedding that fuses image patches with textual names through Q-Former modules.

\textbf{Hybrid Fusion} \cite{16ma2023subject,156zhu2024multibooth}. Moreover, some systems integrate multiple fusion strategies. For example, Subject-Diffusion \cite{16ma2023subject} integrate both concatenation and cross-attention fusion, taking advantage of the strengths of each approach to enhance the overall personalization capability.

\subsubsection{Training Data}
%
Training a PTA model for PCS necessitates a large-scale dataset.
There are primarily two types of training samples utilized:

\textbf{Triplet Data (Reference Image, Target Image, Target Caption).} This dataset format is directly aligned with the PCS objectives, establishing a clear relation between the reference and the personalized content.
However, such large-scale triplet samples are not widely available. Several strategies have been proposed to mitigate this issue:
1) Data Augmentation. Techniques such as foreground segmentation followed by placement in a different background are used to construct triplet data \cite{10li2024blip}.
2) Synthetic Sample Generation. Methods like SuTI \cite{11chen2024subject} utilize multiple TTF models to generate synthetic samples, which are then paired with original references.
3) Utilizing Recognizable SoIs. Collecting images of easily recognizable subjects, such as celebrities, significantly facilitates face personalization \cite{60yuan2023inserting}.

\textbf{Dual Data (Reference Image, Reference Caption).} This dataset is essentially a simplified version of the triplet format, where the personalized content is the original image itself. 
Such datasets are more accessible, including collections like LAION \cite{schuhmann2022laion} and LAION-FACE \cite{zheng2022general}.
However, a notable drawback is that training tends to focus more on reconstructing the reference image rather than incorporating the text prompts. Consequently, models trained on this type of data might struggle with complex prompts that require substantial modifications or interactions with objects.

\subsection{Hybrid Framework}
Recently, some works have started to explore the combination of TTF and PTA methods. 
HyperDreamBooth \cite{58ruiz2023hyperdreambooth} states that PTA methods provide a general framework capable of handling a wide range of common objects, while TTF techniques enable fine-tuning to specific instances, which can improve the preservation of fine-grained details. They first develop a PTA network, followed by a subject-driven fine-tuning.
Similarly, DreamTuner \cite{86hua2023dreamtuner} pre-trains a subject encoder that outputs diffusion conditions for accurate reconstruction. Then an additional fine-tuning stage is conducted for fine identity preservation.
In contrast, SuTI \cite{11chen2024subject} first applies TTF methods to generate synthetic pairwise samples, which can be used for training the PTA network.

\section{Techniques in Personalized Content Synthesis}
\label{sec:techniques}

Building upon the architectural framework discussed in Section \ref{sec:framework}, this section analyzes learning optimization techniques applicable to both frameworks. We focus on four categories: attention mechanisms, mask-guided generation, data augmentation, and regularization strategies.
These methods aim to address key challenges in PCS, like enhancing subject fidelity, minimizing the interference of redundant semantics, enhancing generalization, and avoiding overfitting.

\subsection{Attention-based Operation}
\label{sec:attention}

Attention-based operations have become a crucial technique in model learning, particularly for processing features effectively \cite{hassanin2024visual}. 
In diffusion models, these operations generally involve manipulating the way a model focuses on different parts of data, often through a method known as the Query-Key-Value (QKV) scheme.
%
While large-scale pre-training has equipped this module with strong feature extraction capabilities, there remains significant ongoing work to enhance its performance for customization tasks.

\textbf{Explicit attention weight manipulation}. A cluster of studies focuses on restricting the influence of the SoI token within the attention layers.
%
For example, Mix-of-Show \cite{46gu2024mix} designs region-aware cross-attention where the feature map is initially generated by the global prompt and replaced with distinct regional features corresponding to each entity. This avoids the misalignment between the words and visual regions.
DreamTuner \cite{86hua2023dreamtuner} designs a self-subject-attention layer to further refine the subject identity. This attention module takes the features of the generated image as query, the concatenation of generated features as key, and reference features as value.
%
%
Layout-Control \cite{152chen2024training} adjusts attention weights specifically around the layout without additional training.
Cones 2 \cite{62liu2023cones} also defines some negative attention areas to penalize the illegal occupation to allow multiple object generation.
VICO \cite{20hao2023vico} inserts a new attention layer where a binary mask is deployed to selectively obscure the attention map between the noisy latent and the reference image features.

\textbf{Implicit attention guidance}. In addition to these explicit attention weights modification methods, many researchers \cite{54xiao2023fastcomposer,8han2023svdiff,74avrahami2023break,16ma2023subject,81safaee2023clic,44rahman2024visual,90hyung2023magicapture} employ localization supervision in the cross-attention module.
Specifically, they train cross-attention modules using coordinate-aware loss functions, forcing attention maps to align with annotated subject positions.
DreamTuner \cite{86hua2023dreamtuner} further refines this approach by designing an attention layer that effectively integrates features from different parts of the image.

\subsection{Mask-guided Generation}
\label{sec:mask}
Since reference images contain both the SoI and irrelevant visual elements, masks as a crucial prior indicating the position and contour of the specified object can effectively minimize the influence of redundant information.

\textbf{Pixel-level mask}. Benefitting from advanced segmentation methods like SAM \cite{kirillov2023segment}, the SoI can be precisely isolated from the background.
Based on this strategy, plenty of studies \cite{21zhang2023personalize,48wang2023hifi,74avrahami2023break,81safaee2023clic,51lu2024object,95achlioptas2023stellar, 90hyung2023magicapture,92tang2023retrieving,34li2023generate} choose to discard the pixels of the background area so that the reconstruction loss can focus on the targeted object and exclude irrelevant disturbances.
Another technique \cite{85cai2023decoupled} further adds the masked background reconstruction for better disentanglement.
In addition, the layout indicated by the pixel mask can be incorporated into the attention modules as a supervision signal \cite{54xiao2023fastcomposer,8han2023svdiff,74avrahami2023break,16ma2023subject,81safaee2023clic,44rahman2024visual,90hyung2023magicapture} to adjust the attention's concentration adaptively.
Moreover, the mask can stitch specific feature maps to construct more informative semantic patterns \cite{19wei2023elite,24jia2023taming,16ma2023subject}.
%

\textbf{Feature-level mask}. In addition to the pixel-level manipulation, masks can be extended to feature-level operations. 
DisenBooth \cite{18chen2023disenbooth} defines an identity-irrelevant embedding with a learnable mask. By maximizing the cosine similarity between the identity-preservation embedding and identity-irrelevant embedding, the mask will adaptively exclude the redundant information, and thus the subject appearance can be better preserved.
AnyDoor \cite{14chen2023anydoor} defines a high-frequency mask that stores detailed SoI features as a condition for the image generation process.
Face-Diffuser \cite{52wang2023high} determines the mask through augmentation from the noise predicted by both a pre-trained text-to-image diffusion model and a PTA personalized model. Each model makes its own noise prediction, and the final noise output is a composite created through mask-guided concatenation.

\subsection{Data Augmentation}
Since the optimization of neural networks requires extensive data, existing PCS methods often struggle to capture complete semantic information of the SoI from limited references, resulting in poor images. 
To address this, various data augmentation strategies are employed to enrich the diversity of SoI references.

\textbf{Compositional augmentation}. Some methods enhance data diversity through classical image augmentation like blending and spatial rearrangement.
SVDiff \cite{8han2023svdiff} manually constructs mixed images of multiple SoI as new training data, thereby enhancing the model's exposure to complex scenarios. Such concept composition is also used in other works \cite{16ma2023subject,138chen2023videodreamer,145wang2024customvideo}.
BLIP-Diffusion \cite{10li2024blip} segments the foreground subject and composes it in a random background so that the original text-image pairs are expanded to a larger dataset.
StyleAdapter \cite{26wang2023styleadapter} chooses to shuffle the image patches to break the irrelevant subject and preserve the desired style.
PACGen \cite{34li2023generate} shows that the spatial position entangles with the identity information. Thus rescaling, center crop, and relocation are effective augmentation solutions.

\textbf{Synthetic data}. Generated synthetic data can provide a large amount of training resources when coupled with quality assurance mechanisms.
SuTI \cite{11chen2024subject} establishes a cascaded pipeline where TTF models first generate diverse variations of each SoI. These synthetic samples then train the target PTA model.
Similarly, DreamIdentity \cite{89chen2023dreamidentity} leverages the existing knowledge of celebrities embedded in large-scale pre-trained diffusion model to generate both the source image and the edited face image.
StyleDrop \cite{5sohn2024styledrop} and GAL \cite{151zhang2024generative} implement iterative refinement pipelines where high-quality synthetic outputs from early training phases are incorporated into subsequent rounds.

\textbf{External sources}. It is intuitive to leveraging web resources to expand training datasets. COTI \cite{3yang2023controllable} adopts a scorer network to progressively expand the training set by selecting semantic-relevant samples with high aesthetic quality from a large web-crawled data pool.

\subsection{Regularization}
Regularization is an effective method that is used to regularize the weight update to avoid overfitting and enhance generalization. 


\textbf{Auxiliary data regularization}. To mitigate the overfitting problem where the PCS system consistently produces identical outputs as the references, studies start to use an additional dataset composed of images with the same category of the SoI \cite{2ruiz2023dreambooth}. By reconstructing these images, the personalized model is required to generate diverse instances of the class while adapting to the target subject.
Building on this strategy, StyleBoost \cite{53park2023styleboost} introduces an auxiliary style-specific data to separate content and aesthetic adaptation.
%
Later, a dataset \cite{57he2023data} is curated with detailed textual prompts (specifying attributes/contexts) to improve disentanglement between subject characteristics and background features.

\textbf{Text embedding constraints}. The semantic richness of pre-trained text (e.g., subject class name) provides a powerful regularization signal for personalized generation. 
By strategically constraining how subject-specific representations interact with established linguistic concepts in the embedding space, these approaches can achieve better generalization ability.
%
%
For example, Perfusion \cite{7tewel2023key} constrains key projections toward class noun embeddings while learning value projections from subject images.
Inspired by coached active learning \cite{wei2012coaching,wei2011coached}, which uses anchor concepts for optimization guidance, Compositional Inversion \cite{9zhang2023compositional} employs a set of semantically related tokens as anchors to constrain the token embedding search.
%
%
In addition, some works \cite{12gal2023encoder,51lu2024object} regularize learnable token offsets relative to pre-trained CLIP embeddings. By minimizing the offset, the final word embedding is able to achieve better text alignment.
Similarly, Cones 2 \cite{62liu2023cones} minimizes the offset by reconstructing the features of 1,000 sentences containing the class noun. And \cite{66pang2023cross} optimizes the learnable token towards the mean textual embedding of 691 well-known names.
Domain-Agnostic \cite{17arar2023domain} proposes to use a contrastive loss to guide the SoI text embedding close to its nearest CLIP tokens pre-trained on large-scale samples.
On the other hand, VICO \cite{20hao2023vico} empirically finds that the end-of-text token \texttt{<|EOT|>} keeps the semantic consistency of SoI. To leverage this discovery, an L2 loss is leveraged to reduce the difference of attention similarity logits between the SoI token and \texttt{<|EOT|>}.

\section{Categorization of Image Personalization Tasks}
\label{task}
\begin{figure*}[h]
    \centering
    \includegraphics[width=1\textwidth]{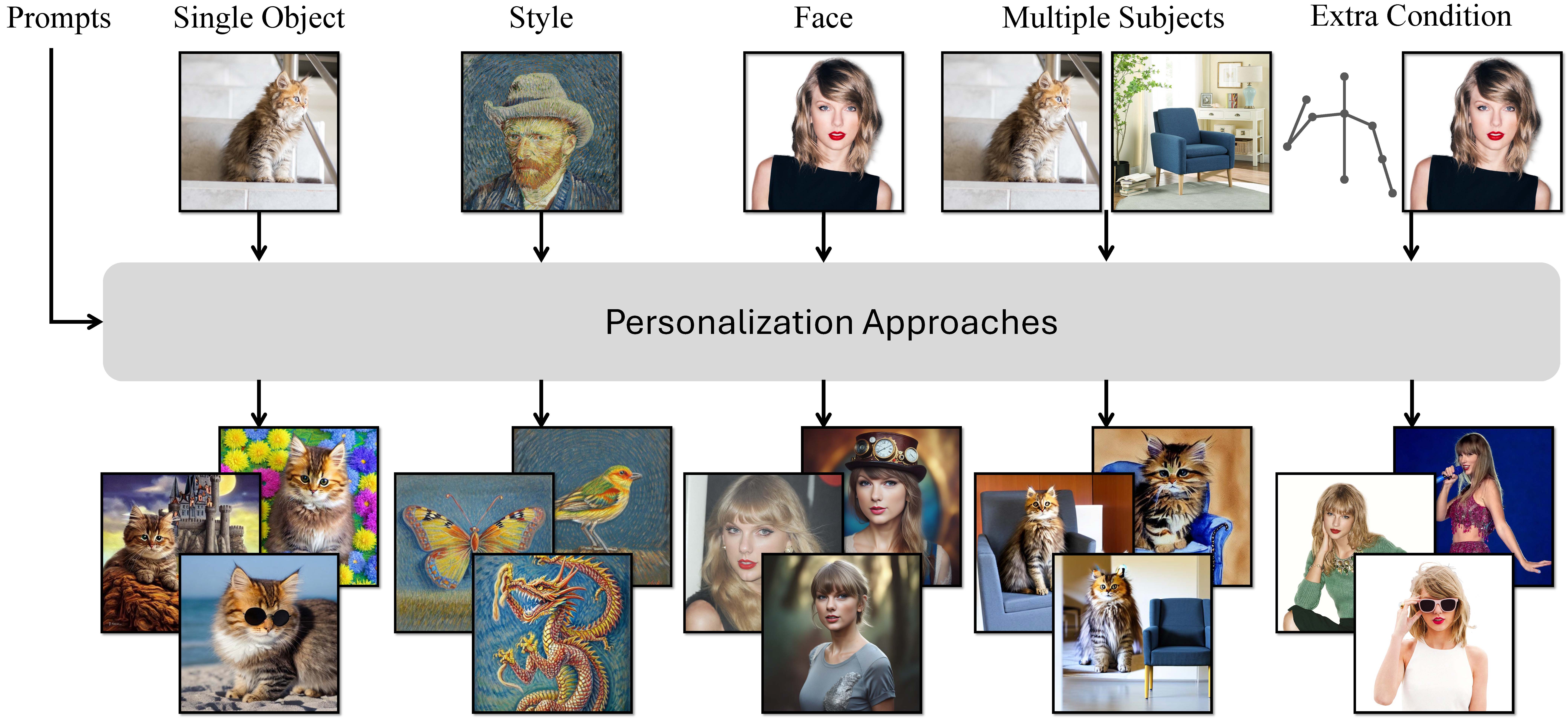}
    \caption{Visualization of several personalized image synthesis tasks. Given text prompts and a few images of the subject of interest, personalization approaches are required to produce expected images.}
    \label{fig:cats}
\end{figure*}

As shown in Fig.~\ref{fig:cats}, personalization covers a range of areas, including objects, styles, faces, etc. 
The following subsections analyze these tasks through the frameworks established in Section \ref{sec:framework}, examining both TTF and PTA approaches for each domain. 
We also summarize these studies in Table \ref{tb:summary1}, Table \ref{tb:summary2}, and Table \ref{tb:summary3} to provide a clear overview and facilitate quick comparison.

\subsection{Personalized Object Generation}
\label{sec:object}
As the foundational task, personalized object generation requires learning discriminative features from general instances (e.g., toys, vehicles, or architecture) and rendering them in novel contexts specified by textual prompts.

\textbf{TTF framework}.
TTF methods perform instance-specific optimization during inference by fine-tuning model parameters on reference images.
This achieves superior subject fidelity and handles rare attributes effectively.

An important branch in TTF methods is to optimize the \textit{learnable token embeddings}. The first work started from Textual Inversion \cite{1gal2022image}, which applies a simple yet effective method that introduces the unique modifier as a new token to represent the SoI.
One of the significant benefits of this method is its minimal storage requirement as the new tokens consumes just a few kilobytes. However, the method compresses complex visual features into a small set of parameters, which can lead to long convergence times and a potential loss in visual fidelity.
Recent work aims to address these limitations through several directions.
DVAR \cite{69voronov2024loss} improves training efficiency by proposing a clear stopping criterion by removing all randomness to indicate the convergence. 
For enhanced representation, P+ \cite{23voynov2023p+} introduces distinct learnable tokens across different layers of the U-Net architecture, thereby offering better attribute control through additional learnable parameters.
NeTI \cite{61alaluf2023neural} advances this concept by proposing a neural mapper that adaptively outputs token embeddings based on the denoising timestep and specific U-Net layers.
ProSpect \cite{30zhang2023prospect} recommends optimizing multiple token embeddings tailored to different denoising timesteps based on the observation that different types of prompts, like layout, color, structure, and texture, are activated at different stages of the denoising process.
Similarly, a study by \cite{79agarwal2023image} shows layered activation insight to learn distinct attributes by selectively activating the tokens within their respective scopes.
Later, HiFiTuner \cite{48wang2023hifi} integrates multiple techniques into the learnable token, including mask-guided loss function, parameter regularization, time-dependent embedding, and generation refinement assisted by the nearest reference.
Alternative approaches, like DreamArtist \cite{68dong2022dreamartist}, choose to optimize both negative and positive prompt embeddings to refine the detail preservation.
%
%
In addition to these token-level refinement approaches, the field continues to evolve with novel techniques such as InstructBooth's reinforcement learning framework \cite{56chae2023instructbooth} and gradient-free evolutionary optimization \cite{73fei2023gradient}.
%
%
In summary, recent developments following the foundational work of Textual Inversion focus on reducing training times and enhancing the visual quality of generated images.

In the realm of TTF methods for PCS, there is a clear shift towards \textit{fine-tuning model weights} rather than just the token embeddings. This approach often addresses the limitations where token embeddings alone struggle to capture complex semantics uncovered in the pre-training data \cite{46gu2024mix,80zhao2023catversion}.
DreamBooth \cite{2ruiz2023dreambooth} proposes to use a unique modifier by a rare token to represent the SoI and fine-tune the whole parameters of the diffusion model. Besides, a regularization dataset containing 20-30 images with the same category as SoI is adopted to overcome the overfitting problem. These two combined approaches achieve impressive performance that largely promotes the progress of the research on image personalization.
However, fine-tuning the entire model for each new object causes considerable storage costs, potentially rendering widespread application. 
To address this, Custom Diffusion \cite{6kumari2023multi} focuses on identifying and fine-tuning critical parameters, particularly the key-value projections in cross-attention layers, to achieve a balance of visual fidelity and storage efficiency.
Further approach, Perfusion \cite{7tewel2023key}, also adopts the cross-attention fine-tuning and proposes to regularize the update direction of the K (key) projection towards the super-category token embedding and the V (value) projection towards the learnable token embedding. 
%
COMCAT \cite{75xiao2023comcat} introduces a low-rank approximation of attention matrices, which drastically reduces storage requirements to 6 MB while maintaining high fidelity in the outputs.
Additionally, methods like adapters \cite{72xiang2023closer,5sohn2024styledrop} and LoRA variants \cite{46gu2024mix,58ruiz2023hyperdreambooth,115kong2024omg,51lu2024object,95achlioptas2023stellar,76yeh2023navigating} are increasingly utilized in personalized generation for parameter-efficient fine-tuning.
It is worth noting that the token embedding fine-tuning is compatible with diffusion weight fine-tuning. 
%
Multiple methods \cite{90hyung2023magicapture,74avrahami2023break,86hua2023dreamtuner} have started using combined weight fine-tuning. 

\textbf{PTA framework}.
For practical deployment of PCS systems, fast response time is a crucial factor. 
PTA methods enable real-time generation (less than 10 sec/subject) by leveraging large-scale pre-training to avoid per-subject optimization during the inference phase.
Re-Imagen \cite{67chen2022re} introduces a retrieval-augmented generative approach, which leverages features from text-image pairs retrieved via a specific prompt. While it is not specifically tailored for object personalization, it demonstrates the feasibility of training reference-conditioned frameworks.
Later, ELITE \cite{19wei2023elite} specifically targets image personalization by combining the global reference features with text embedding while incorporating local features that exclude irrelevant backgrounds. Both fused features and local features serve as conditions for the denoising process.
Similarly, InstantBooth \cite{22shi2023instantbooth} retrains CLIP models to extract image features and patch features, which are injected into the diffusion model via the attention mechanism and learnable adapter, respectively. 
Additionally, UMM-Diffusion \cite{71ma2023unified} designs a multi-modal encoder that produces fused features based on the reference image and text prompt. The text features and multi-modal hidden state are seen as guidance signals to predict a mixed noise.
Another work, SuTI \cite{11chen2024subject}, adopts the same architecture as Re-Imagen. The difference lies in the training samples which are produced by a massive number of TTF models, each tuned on a particular subject set. This strategy promotes a more precise alignment with personalization at an instance level rather than the class level of Re-Imagen.
Moreover, Domain-Agnostic \cite{17arar2023domain} combines a contrastive-based regularization technique to push the pseudo embedding produced by the image encoder towards the existing nearest pre-trained token. Besides, They introduce a dual-path attention module separately conditioned on the nearest token and pseudo embedding. 
Compared to the methods that use separate encoders to process a single modality, some works have explored the usage of pre-trained multi-modal large language models (MLLM) that can process text and image modality within a unified framework.
For example, BLIP-Diffusion \cite{10li2024blip} utilizes the pre-trained BLIP2 \cite{li2023blip} that encodes multimodal inputs including the SoI reference and a class noun. The output embedding is then concatenated with context description and serves as a condition to generate images.
Further, Customization Assistant \cite{84zhou2023customization} and KOSMOS-G \cite{77pan2023kosmos} replace the text encoder of Stable Diffusion with a pre-trained MLLM to output a fused feature based on the reference and context description. Meanwhile, to meet the standard format of Stable Diffusion, a network is trained to align the dimension of the output embedding.

\subsection{Personalized Style Generation}
Personalized style generation seeks to tailor the aesthetic elements of reference images. The concept of ``style" now includes a wide range of artistic elements, such as brush strokes, material textures, color schemes, structural forms, lighting techniques, and cultural influences.

\textbf{TTF framework}. In this field, StyleDrop \cite{5sohn2024styledrop} leverages adapter tuning to efficiently capture the style from a single reference image. This method demonstrates the effectiveness through iterative training, utilizing synthesized images refined by feedback mechanisms, like human evaluations and CLIP scores. This approach not only enhances style learning but also ensures that the generated styles align closely with human aesthetic judgments.
Later, GAL \cite{151zhang2024generative} proposes an uncertainty-based evaluation strategy to filter high-quality synthetic-style data and uses a weighted schema to balance the contribution of the additional samples and the original reference. 
Furthermore, StyleAligned \cite{98hertz2023style} focuses on maintaining stylistic consistency across a batch of images. This is achieved by using the first image as a reference, which acts as an additional key and value in the self-attention layers, ensuring that all subsequent images in the batch adhere to the same stylistic guidelines.
Style-friendly \cite{211choi2024style} introduces a novel diffusion model fine-tuning approach that enhances personalized artistic style generation by adaptively biasing noise sampling toward higher noise levels, where stylistic features emerge.

\textbf{PTA framework}. For the PTA framework, StyleAdapter \cite{26wang2023styleadapter} employs a dual-path cross-attention mechanism within the PTA framework. This model introduces a specialized embedding module designed to extract and integrate global features from multiple style references. 
Diptych Prompting \cite{209shin2024large} utilizes an inpainting mechanism to draw another image with the same style of the reference part.

\subsection{Personalized Face Generation}
Personalized face generation aims to generate diverse identity images that adhere to text prompt specifications, utilizing only a few initial face images. Compared to general object personalization, the scope is narrowed to a specific class, humans. An obvious benefit is that one can readily leverage large-scale human-centric datasets \cite{zheng2022general,yang2020facescape,cao2018vggface2} and utilize pre-trained models in well-developed areas, like face landmark detection \cite{wu2019facial} and face recognition \cite{adjabi2020past}.

\textbf{TTF framework}. Regarding TTF methods, PromptNet \cite{55zhou2023enhancing} trains a diffusion-based network that encodes the input image and noisy latent to a word embedding. To alleviate the overfitting problem, the noises predicted by the word embedding and context description are balanced through fusion sampling in classifier-free guidance.  
%
%
Additionally, Celeb Basis \cite{60yuan2023inserting} provides a novel idea that the personalized ID can be viewed as the composition of celebrity's face, which has been learned by the pre-trained diffusion model. Based on this hypothesis, a simple MLP is optimized at test time to transform face features into the weighting of different celebrity name embeddings.

\textbf{PTA framework}. Thanks to the abundance of available datasets featuring the same individual in various contexts, which provide valuable data for pre-training PTA methods, the number of works in PTA frameworks is rapidly increasing.
Face0 \cite{88valevski2023face0} crops the face region to extract refined embeddings and concatenates them with text features. During the sampling phase, the output of classifier-free guidance is replaced by a weighted combination of the noise patterns predicted by face-only embedding, text-only embedding, and concatenated face-text embedding. 
The $\mathcal{W}+$ Adapter \cite{91li2023stylegan} constructs a mapping network and residual cross-attention modules to transform the facial features from the StyleGAN \cite{karras2019style} $\mathcal{W}+$ space into the text embedding space of Stable Diffusion.
FaceStudio \cite{31yan2023facestudio} adapts the cross-attention layer to support hybrid guidance including stylized images, facial images, and textual prompts.
Moreover, PhotoMaker \cite{28li2023photomaker} constructs a high-quality dataset through a meticulous data collection and filtering pipeline. They use a two-layer MLP to fuse ID features and class embeddings for an overall representation of human portrait. 
%
PortraitBooth \cite{96peng2023portraitbooth} also employs a simple MLP, which fuses the text condition and shallow features of a pre-trained face recognition model. To ensure expression manipulation and facial fidelity, they add another expression token and incorporate the identity preservation loss and mask-based cross-attention loss.
InstantID \cite{37wang2024instantid} additionally introduces a variant of ControlNet that takes facial landmarks as input, providing stronger guiding signals compared to the methods that solely rely on attention fusion.

\subsection{Multiple Subject Composition}
Multiple subject composition refers to the scenario where users intend to compose multiple SoI displayed in one or more references.

\textbf{TTF framework}.
This task presents a challenge for the TTF methods, particularly in how to integrate the parameters within the same module which are separately fine-tuned for individual SoI.
Some works focus on the one-for-one generation following a fusion mechanism.
For instance, Custom Diffusion \cite{6kumari2023multi} proposes a constrained optimization method to merge the cross-attention key-value projection weights with the goal of maximizing reconstruction performance for each subject. 
Mix-of-Show \cite{46gu2024mix} fuses the LoRA \cite{hu2021lora} weights with the same optimization objective.
StyleDrop \cite{5sohn2024styledrop} dynamically summarizes noise predictions from each personalized diffusion model.
%
In OMG \cite{115kong2024omg}, the latent predicted by each LoRA-tuned model is spatially composited using the subject mask.
Joint training is another strategy to cover all expected subjects.
SVDiff \cite{8han2023svdiff} employs a data augmentation method called Cut-Mix to compose several subjects together and applies a location loss to regularize attention maps, ensuring alignment between each subject and its corresponding token.
Similar strategies are found in other works \cite{6kumari2023multi,60yuan2023inserting} which train a single model by reconstructing the appearance of every SoI.
There are advanced control mechanisms designed to manage multiple subjects. 
Cones \cite{4liu2023cones} proposes to find a small cluster of neurons that preserve the most information about SoI. The neurons belonging to different SoI will be simultaneously activated to generate the combination.
Compositional Inversion \cite{9zhang2023compositional} introduces spatial region assignment to different subjects to improve the composition success rate.

\textbf{PTA framework}. For PTA frameworks, multi-subject generation is achieved through specialized architectural designs.
Fastcomposer \cite{54xiao2023fastcomposer}, Subject-Diffusion \cite{16ma2023subject}, and $\lambda$-eclipse \cite{222patel2024lambda} place each subject feature in its corresponding placeholder in the text embedding, ensuring a seamless and efficient combination.
CustomNet \cite{29yuan2023customnet} and MIGC \cite{174zhou2024migc} train a PTA network that supports location control for each subject.
SSR-Encoder \cite{221zhang2024ssr} implements an encoder to selectively preserve the desired subject feature and a cross-attention module to support multi-subject feature fusion.

\subsection{High-level Semantic Personalization}
Recently, the field of image personalization has started to include complex semantic relationships and high-level concepts. Different approaches have been developed to enhance the capability of models to understand and manipulate these abstract elements. 

\textbf{TTF framework}. For now, all researches in this field are based on the TTF framework.
ReVersion \cite{63huang2023reversion} intends to invert object relations from references. Specifically, they use a contrastive loss to guide the optimization of the token embedding towards specific clusters of Part-of-Speech tags, such as prepositions, nouns, and verbs. Meanwhile, they also increase the likelihood of adding noise at larger timesteps during the training process to emphasize the extraction of high-level semantic features.
Lego \cite{45motamed2023lego} focuses on more general concepts, such as adjectives, which are frequently intertwined with the subject appearance. The concept can be learned from a contrastive loss applied to the dataset comprising clean subject images and images that embody the desired adjectives.
Moreover, ADI \cite{43huang2023learning} aims to learn the action-specific identifier from the references. To ensure the inversion only focuses on the desired action, ADI extracts gradient invariance from a constructed triplet sample and applies a threshold to mask out the irrelevant feature channels.
%

\section{Extensions of Personalized Content Synthesis}
\label{sec:extension}
While the core PCS system focuses on image generation from reference subjects, recent advances have expanded its capabilities across multiple dimensions. This section examines several frontier extensions that push the boundaries of personalization technology.

\subsection{Personalization on Extra Conditions}
Recent personalization tasks tend to include additional conditions for diverse content customization.
One common application is to customize the subject into a fixed source image.
%
For example, PhotoSwap \cite{15gu2024photoswap} introduces a new task that replaces the subject in the source image with the SoI from the reference image. To address this requirement, they first fine-tunes a diffusion model on the references to obtain a personalized model. To preserve the background of the source image, they initialize the noise with DDIM inversion \cite{song2020denoising} and replace the intermediate feature maps with those derived from source image generation during inference time.
Later, MagiCapture \cite{91li2023stylegan} broadens the scope to face customization.
Another similar application can be found in Virtual Try-on, which aims to fit selected clothing onto a target person. The complexities of this task have been thoroughly analyzed in another survey \cite{islam2024deep}.

Additional conditions in personalization tasks may include adjusting the layout \cite{152chen2024training}, transforming sketches \cite{94chen2023democaricature}, controlling viewpoint \cite{29yuan2023customnet,163kumari2024customizing}, or modifying poses \cite{37wang2024instantid}. Each of these conditions presents unique challenges and requires specialized approaches to integrate these elements seamlessly into the personalized content.

\subsection{Personalized Video Generation}
With the rising popularity of video generation \cite{xing2024survey}, video personalization has also begun to attract attention. 
In video personalization, the SoI can be classified into three distinct categories: appearance, motion, and the combination of both appearance and motion.

\textbf{Appearance-based video personalization}. This task focuses on transferring subject appearance from static images to video sequences. The standard TTF pipeline utilizes reference images as appearance anchors and fine-tunes video diffusion models (VDM) for temporal synthesis. 
The process involves leveraging sophisticated methods from 2D personalization, such as parameter-efficient fine-tuning \cite{145wang2024customvideo}, data augmentation \cite{138chen2023videodreamer,145wang2024customvideo}, and attention manipulation \cite{83jiang2023videobooth,212liu2024stylecrafter,145wang2024customvideo}. 
Additionally, several studies \cite{82zhao2023videoassembler,83jiang2023videobooth,212liu2024stylecrafter,105feng2023dreamoving} have explored the PTA framework. 
These diffusion models are specifically tailored to synthesize videos based on the image references.

\textbf{Motion-based video personalization}. In this task, the reference input switches to a video clip containing a consistent action. 
A common approach is to fine-tune the video diffusion model by reconstructing the action clip \cite{150wu2023tune,103song2023save,140jeong2023vmc,148esser2023structure,136he2023animate,102wu2023lamp,104materzynska2023customizing}. 
However, distinguishing between appearance and motion within the reference video can be challenging. 
To solve this problem, SAVE \cite{103song2023save} applies appearance learning to ensure that appearance is excluded from the motion learning phase. 
Additionally, VMC \cite{140jeong2023vmc} removes the background information during training prompt construction.

\textbf{Appearance and motion personalization}. When integrating both subject appearance and motion, innovative methods are employed to address the complexities of learning both aspects simultaneously. 
MotionDirector \cite{101zhao2023motiondirector} utilizes spatial and temporal losses to facilitate learning across these dimensions. 
Another approach, DreamVideo \cite{142wei2023dreamvideo}, incorporates residual features from a randomly selected frame to emphasize subject information. This technique enables the fine-tuned module to primarily focus on learning motion dynamics.

In summary, video personalization strategies vary significantly based on the specific aspects. Moreover, due to the current limitations in robust video feature representation, PTA video personalization that is directly conditioned on video input remains an area under exploration.

\subsection{Personalized 3D generation}
Personalized 3D Generation refers to the process of creating customized 3D models or scenes based on 2D SoI images.
Basically, the pipeline begins by fine-tuning a 2D diffusion model using TTF methods. This tuned model then utilize Score Distillation Sampling (SDS) \cite{poole2022dreamfusion} to train a 3D Neural Radiance Field (NeRF) model \cite{mildenhall2021nerf} for each specific prompt \cite{120lin2023magic3d,126shi2023mvdream}.
Building on this foundation, several methods have been developed to improve the workflow.
DreamBooth3D \cite{121raj2023dreambooth3d} structures the process into three phases: initializing and optimizing a NeRF from a DreamBooth model, rendering multi-view images, and fine-tuning a secondary DreamBooth for the final 3D NeRF refinement.
Consist3D \cite{127ouyang2023chasing} enhances text embeddings by training two distinct tokens, a semantic and a geometric token, during 3D model optimization.
TextureDreamer \cite{133yeh2024texturedreamer} focuses on extracting texture maps from optimized spatially-varying bidirectional reflectance distribution (BRDF) fields for rendering texture on wide-range 3D subjects.

Additionally, advancements extend to 3D avatar rendering and dynamic scenes.
Animate124 \cite{128zhao2023animate124} and Dream-in-4D \cite{130zheng2023unified} integrate video diffusion for 4D dynamic scene support within the 3D optimization process.
In avatar rendering, PAS \cite{122azadi2023text} generates 3D body poses configurable by avatar settings, StyleAvatar3D \cite{123zhang2023styleavatar3d} facilitates 3D avatar generation based on images, and AvatarBooth \cite{124zeng2023avatarbooth} employs dual fine-tuned diffusion models for separate face and body generation.

\subsection{Attack and Defense}
This rapid advancement raises concerns about the ethical implications of PCS, particularly in areas such as misinformation, privacy violations, and the creation of deepfakes. There is an increased risk that individuals or organizations may exploit them to produce misleading content or manipulate public perception.
To mitigate this, Anti-DreamBooth \cite{13van2023anti} aims to add a subtle noise perturbation to the references so that any personalized model trained on these samples only produces terrible results. The basic idea is to maximize the reconstruction loss of the surrogate model.
Additionally, \cite{50zhang2023backdooring} suggests to predefine a collection of trigger words and meaningless images. These data are paired and incorporated during the training phase. Once the trigger words are encountered, the synthesized image will be intentionally altered for safeguarding.

\subsection{Other Emerging Directions}
Several works are exploring different personalization extensions.
For example, SVG personalization is introduced by \cite{36zhang2023text}, in which a parameter-efficient fine-tuning method is applied to create SVGs. After first-step generation, the SVGs are refined through a process that includes semantic alignment and a dual optimization approach, which utilizes both image-level and vector-level losses to enhance the final output.
%
%
Another application, 360-degree panorama customization \cite{25wang2024customizing}, is also emerging as a potential tool for personalization in the digital imaging realm.

\section{Evaluation}
\label{evaluation}

\subsection{Evaluation Dataset} 
%
To assess the performance of personalized models, various datasets have been developed:

\textbf{DreamBench} \cite{2ruiz2023dreambooth} serves as the primary evaluation benchmark for DreamBooth \cite{2ruiz2023dreambooth}, containing 30 diverse subjects (e.g., backpacks, animals, vehicles, and toys) with 25 unique prompts per subject.

\textbf{DreamBench-v2} \cite{11chen2024subject} expands the evaluation scope of DreamBench by adding 220 test prompts for each subject.

\textbf{Custom-10} \cite{6kumari2023multi} used in Custom Diffusion \cite{6kumari2023multi} evaluates 10 subjects, each with 20 specific test prompts, and includes tests for multi-subject composition with 5 pairs of subjects and 8 prompts for each pair.

\textbf{Custom-101} \cite{6kumari2023multi} is the latest released dataset by the authors of Custom Diffusion \cite{6kumari2023multi}, which comprises 101 subjects to provide a broader scope of evaluation. 

\textbf{Stellar} \cite{95achlioptas2023stellar} specifically targets human-centric evaluation, featuring 20,000 prompts on 400 human identities.

Despite these contributions, they remain fragmented across different research groups and the research community still lacks a benchmark tested on a large number of personalized generation tasks. To address this gap, this survey introduces a comprehensive evaluation dataset\footnote{This dataset is available on: \url{https://github.com/zhangxulu1996/awesome-personalization}} designed for the most common personalized object and face personalization. Section\ref{sec:benchmark} details our benchmark design and evaluation results.


\subsection{Evaluation Metrics}

As PCS aims to maintain fidelity to the SoI while ensuring alignment with textual conditions, the metrics are designed from two aspects, text alignment and visual fidelity. 

The text alignment metrics quantify how precisely generated outputs reflect prompt semantics:

\textbf{CLIP-T} measures semantic alignment using the cosine similarity between CLIP \cite{radford2021learning} embeddings of generated images and their text prompts.

\textbf{ImageReward} \cite{xu2023imagereward}, \textbf{HPS Score} (v1/v2) \cite{wu2023human,wu2023humanv2}, and \textbf{PickScore} \cite{kirstain2023pick} employ learned models trained on human judgments to better correlate with perceptual quality to reflect human preference better.

To determine how closely the generated subject resembles the SoI, visual fidelity can be assessed via the following metrics:

\textbf{CLIP-I} evaluates subject preservation through CLIP image embedding similarity between generations and references. Optimal values should balance fidelity (high scores) against overfitting (excessive scores that ignore text guidance).

\textbf{DINO-I} \cite{caron2021emerging} provides a complementary assessment using DINO's instance-aware features, particularly effective for object-level similarity.

\textbf{Fréchet Inception Distance (FID)} \cite{FIDheusel2017gans} quantifies the statistical similarity between generated and real image distributions through Inception-V3 \cite{inception}. 


In addition to these commonly adopted metrics, there are some discussions of specialized metrics for PCS system evaluation:
LyCORIS \cite{76yeh2023navigating} introduces a 5-dimensional assessment covering fidelity, controllability, diversity, base model preservation, and image quality.
Stellar \cite{95achlioptas2023stellar} develops six human-centric metrics including soft-penalized CLIP text score, Identity Preservation Score, Attribute Preservation Score, Stability of Identity Score, Grounding Objects Accuracy, and Relation Fidelity Score. These metrics ensure a structured and detailed evaluation of personalized models.
This evolving metric landscape reflects the growing sophistication of PCS systems, with optimal evaluation typically requiring combinations of general and task-specific measures.

\subsection{New Benchmark on SoTA Methods}
\label{sec:benchmark}
Although multiple evaluation datasets have been proposed in this area, there is still an urgent need for a standardized benchmark that systematically assesses performance across diverse PCS methodologies. To address this limitation, we present a new test dataset called \textbf{Persona} with a comprehensive evaluation built upon existing works.

\textbf{For Object.} Persona includes 47 subjects from available resources \cite{1gal2022image, 2ruiz2023dreambooth, 6kumari2023multi}.
%
Following the methodology of DreamBooth \cite{2ruiz2023dreambooth}, we categorize the subjects into two classes: objects and live pets, based on whether the subject is a living entity. Specifically, 10 out of 47 subjects are pets and the remaining 37 are various objects.
%
To evaluate the performance, we utilize the text prompts from DreamBooth \cite{2ruiz2023dreambooth}. This includes 20 recontextualization prompts and 5 property modification prompts for objects, along with 10 recontextualization, 10 accessorization, and 5 property modification prompts for pets, totally 25 prompts per category.

\textbf{For Face.} We also collected 15 subjects from Celeb-A \cite{liu2018large} into our Persona dataset. We use 40 prompts for evaluation, including 10 accessory prompts, 10 style prompts, 10 action prompts, and 10 context prompts.

\textbf{Settings}. To assess existing representative PCS methods using our constructed test dataset, we compute CLIP-T to assess the text alignment.
We calculate CLIP-I for subject fidelity in object generation.
For subject fidelity in face generation, we detect faces in both the generated and target images using MTCNN \cite{zhang2016mtcnn} and calculate the pairwise identity similarity using FaceNet \cite{schroff2015facenet}.
To uniformly evaluate and compare existing personalized generation methods, we select 22 representative PCS methods for evaluation. 
%
%
%
We generate 4 images for each test prompt and set the same random seed for all methods.

\textbf{Results}. The evaluation results are shown in Table \ref{tb:eval}. 
It is evident that no method excels simultaneously in both visual fidelity and text alignment metrics. This highlights a significant challenge currently faced by PCS methods: achieving the optimal trade-off between subject preservation and editability. 
Striking this balance is difficult, as high subject fidelity often comes at the cost of prompt fidelity and vice versa.
Furthermore, we note that higher visual fidelity does not always equate to better performance. The generated images sometimes exhibit patterns that closely mirror the reference images and ignore the prompt guidance.
This phenomenon primarily arises from model overfitting on the reference input, which hinders the model's ability to generalize. 
Consequently, the visual similarity metric yields higher results based on the mirrored output and the reference, rather than providing an accurate representation of the performance.

\begin{table*}[t]
    \centering
    \caption{Evaluation results of representative methods on our \textbf{Persona} evaluation dataset.}
    \begin{small}
    \begin{tabular*}{\textwidth}{@{\extracolsep{\fill}} l|l|c|c|cc @{}}
        \hline
         Type &Methods &Framework & Backbone & CLIP-T & CLIP-I \\
        \hline
        \multirow{15}{*}{Object} 
         & Textual Inversion \cite{1gal2022image} & TTF  & SD 1.5 & 0.199 & 0.749 \\
         & Dreambooth \cite{2ruiz2023dreambooth} & TTF & SD 1.5 & 0.286 & 0.772 \\
         & P+ \cite{23voynov2023p+}  & TTF & SD 1.4& 0.244  & 0.643 \\
         & Custom Diffusion \cite{6kumari2023multi}  & TTF & SD 1.4 & 0.307 & 0.722 \\
         & NeTI \cite{61alaluf2023neural}  & TTF &SD 1.4  &0.283 & 0.801 \\
         & SVDiff \cite{8han2023svdiff}  & TTF & SD 1.5& 0.282  & 0.776 \\
         & Perfusion \cite{7tewel2023key} & TTF & SD 1.5& 0.273  & 0.691 \\`
         & ELITE \cite{19wei2023elite} & PTA & SD 1.4 &0.292&0.765 \\
         & BLIP-Diffusion \cite{10li2024blip} & PTA & SD 1.5 & 0.292 & 0.772 \\
         & IP-Adapter \cite{40ye2023ip}  & PTA & SD 1.5 & 0.2722  & \textbf{0.825} \\
         & SSR Encoder \cite{221zhang2024ssr}  & PTA & SD 1.5 & 0.288 & 0.792 \\
         & MoMA \cite{song2024moma} & PTA & SD 1.5 & 0.322 &0.748 \\
         & Diptych Prompting \cite{209shin2024large} & PTA & FLUX 1.0 dev & \textbf{0.327} & 0.722\\
         & $\lambda$-eclipse \cite{222patel2024lambda} & PTA &Kandinsky 2.2 &0.272 & 0.824\\
         & MS-Diffusion \cite{220personalizationms} & PTA & SDXL & 0.298 &0.777 \\
        \hline
        \multirow{8}{*}{Face} 
        & CrossInitialization \cite{66pang2023cross} &TTF & SD 2.1 &0.261 &0.469 \\
        & Face2Diffusion \cite{170shiohara2024face2diffusion}  & PTA & SD 1.4 &0.265 &0.588 \\
         & SSR Encoder \cite{221zhang2024ssr} & PTA & SD 1.5 &0.233 &0.490 \\
        & FastComposer \cite{54xiao2023fastcomposer} & PTA & SD 1.5 & 0.230 & 0.516 \\
         & IP-Adapter \cite{40ye2023ip} & PTA & SD 1.5 &0.292 &0.462 \\
         & IP-Adapter \cite{40ye2023ip} & PTA & SDXL &0.292 &0.642 \\
         & PhotoMaker \cite{28li2023photomaker} & PTA & SDXL & \textbf{0.311} & 0.547 \\
         & InstantID \cite{37wang2024instantid}  & PTA & SDXL & 0.278 & \textbf{0.707} \\
         \hline
    \end{tabular*}
    \end{small}
    \label{tb:eval}
\end{table*}

\section{Challenge and Outlook}
\label{challenge}
\subsection{Overfitting Problem} 
As discussed in Section \ref{sec:benchmark}, current PCS systems face a critical challenge of overfitting because of a limited set of reference images. 
This overfitting problem manifests in two ways: 
1) Loss of SoI editability. The personalized model tends to produce images that rigidly mirror the SoI in the reference, such as consistently depicting a cat in an identical pose.
2) Irrelevant semantic inclusion. The irrelevant elements in the references are generated in the output, such as backgrounds or objects not pertinent to the current context.

To investigate the rationale behind, Compositional Inversion \cite{9zhang2023compositional} observes that the learned token embedding is located in an out-of-distribution area compared to the center distribution formed by pre-trained words. 
This is also found in another work \cite{66pang2023cross} that the learnable token embeddings deviate significantly from the distribution of the initial embedding. 
In addition, there is evidence \cite{9zhang2023compositional,24jia2023taming,20hao2023vico} suggesting that the unique modifier dominates in the cross-attention layers compared to the other context tokens, leading to the absence of other semantic appearances.

To address this issue, many solutions have been proposed. Most methods discussed in Section \ref{sec:techniques} contribute to the alleviation of the overfitting problem, such as the exclusion of redundant background, attention manipulation, regularization of the learnable parameters, and data augmentation. 
However, it has not been solved yet, especially in the cases where the SoI has a non-rigid appearance \cite{9zhang2023compositional} or the context prompt has a similar semantic correlation with the irrelevant elements in the reference \cite{151zhang2024generative}.
It is clear that addressing overfitting in PCS is not merely a technical challenge but a necessity for ensuring the practical deployment and scalability of these systems in varied and dynamic real-world environments.
Therefore, an effective strategy and robust evaluation metrics are urgently needed to achieve broader adoption and greater satisfaction in practical applications.

\subsection{Trade-off on Subject Fidelity and Text Alignment} 
The ultimate goal of personalized content synthesis is to create systems that not only render the SoI with high fidelity but also effectively respond to textual prompts.
However, achieving excellence in both areas simultaneously presents a notable conflict.
Specifically, high subject fidelity typically involves capturing and reproducing detailed and specific features of the SoI. This often requires the model to minimize the reconstruction loss to replicate delicate characteristics accurately.
Conversely, text alignment necessitates that the system flexibly adapts the SoI according to varying textual descriptions. These descriptions may suggest changes in pose, expression, environment, or stylistic alterations that do not aim to reconstruct the exact visualization in the reference.
As a result, it becomes challenging to achieve flexible adaptation in different contexts while simultaneously pushing the model to capture fine-grained details.
%
%
To address this inherent conflict, Perfusion \cite{7tewel2023key} proposes to regularize the attention projections by these two items.
\cite{35cao2023concept} decouples the conditional guidance into two separate processes, which allows for the distinct handling of subject fidelity and textual alignment.
Despite these efforts, there still remains room for further exploration and refinement of this issue. Enhanced model architectures, innovative training methodologies, and more dynamic data handling strategies could potentially provide new pathways to better balance the demands of subject and text fidelity in PCS systems.


\subsection{Standardization and Evaluation} 

Despite the popularity of personalization, there is a noticeable lack of standardized test datasets and robust evaluation metrics that accurately capture the performance of different strategies.
Currently, a widely used metric for assessing visual fidelity relies on CLIP image similarity. However, this approach may wrongly exaggerate the value when the model is overfitted to the references.
Therefore, future efforts should focus on creating comprehensive and widely accepted benchmarks that can evaluate various aspects of PCS models, including but not limited to visual fidelity and subject editability.

\subsection{Multimodal Autoregressive Frameworks}
Recent advancements in multimodal autoregressive models present novel solutions for PCS by unifying cross-modal understanding and generation. Models like Emu3 \cite{wang2024emu3} demonstrate that autoregressive architectures can natively handle image-text-video sequences through discrete tokenization and joint transformer training. This paradigm enables seamless integration of user-provided multimodal references (e.g., text descriptions paired with SoI images) while maintaining contextual coherence across generation steps. Besides, this framework natively support for subject editing via multi-round chat, effectively addressing the overfitting limitations commonly observed in diffusion-based models.

\subsection{Interactive Personalization Workflow}
The evolution of interactive generation systems has unlocked new frontiers for PCS, particularly through the integration of multi-round interactive generation. This capability allows users to iteratively refine and accurately define the SoI, addressing the challenge of translating vague or complex requirements into precise content generation. For instance, conversational PCS systems like Gemini-2.0-flash \cite{team2023gemini} exemplify this progress, leveraging natural language dialogue to iteratively optimize both subject fidelity and prompt alignment. By enabling users to provide real-time feedback and adjust parameters through chat-like interactions, these systems bridge the gap between abstract intent and concrete outputs, aligning with PCS’s core objective of balancing faithful subject representation with flexible editability.

\section{Conclusion}
This survey has provided a thorough review of personalized content synthesis with diffusion models, particularly focusing on 2D image customization. 
We explore two main frameworks, TTF and PTA methods, and delve into their mechanics. 
We also cover the recent progress in specific customization areas, including object, face, style, video, 3D synthesis.
%
%
In addition to the impressive techniques, we propose several challenges that still need to be addressed. These include preventing overfitting, finding the right balance between reconstruction quality and editability, and standardizing evaluation methods. 
To support ongoing research, we collect a test dataset from existing literature and evaluate the classical method to provide a clear comparison.
By providing detailed analysis and outlining targeted recommendations, we hope to promote further innovation and collaboration within the PCS community.

\FloatBarrier
\begin{table*}[htb]
    \centering 
    \tiny
    \caption{Paper summary on personalization.} 
    \label{tb:summary1}
\begin{tabularx}{\textwidth}{@{}| X |>{\centering\arraybackslash}p{0.14\textwidth} |>{\centering\arraybackslash}p{0.1\textwidth} |>{\centering\arraybackslash}p{0.13\textwidth} |p{0.28\textwidth}  |@{}}
    \hline
 \textbf{Paper} & \textbf{Scope} & \textbf{Framework} &\textbf{Backbone} & \textbf{Technique} \\ \hline
\rowcolor{c1}     Textual Inversion \cite{1gal2022image} & Object  & TTF        & LDM             & Token Embedding Fine-tuning                                                      \\\hline
\rowcolor{c1}   DreamBooth \cite{2ruiz2023dreambooth}           & Object  & TTF              & Imagen         & \makecell[l]{Diffusion Model Fine-tuning; \\Regularization Dataset}                                                        \\\hline
\rowcolor{c1}   DreamArtist \cite{68dong2022dreamartist}         & Object  & TTF               & LDM          & Negative Prompt Fine-tuning                                                  \\\hline
\rowcolor{c1}   DVAR \cite{69voronov2024loss}                    & Object  & TTF             & SD 1.5         & Randomness Erasing                                             \\\hline
\rowcolor{c1}   HiPer \cite{33han2023highly}                     & Object  & TTF              & SD      & Token Embedding Enhancement                                          \\\hline
\rowcolor{c1}   P+ \cite{23voynov2023p+}  & Object  & TTF            & SD 1.4                  & Token Embedding Enhancement                            \\\hline
\rowcolor{c1}   Unet-finetune \cite{72xiang2023closer}           & Object  & TTF               & SD          & Parameter-efficient Fine-tuning                                 \\\hline
\rowcolor{c1} Jia et al. \cite{24jia2023taming}                 & Object  & TTF           & Imagen           & \makecell[l]{Mask-assisted Generation; \\ Regularization }                                           \\\hline
\rowcolor{c1} COTI    \cite{3yang2023controllable}            & Object  & TTF              & SD 2.0         & Data Augmentation                                             \\\hline
\rowcolor{c1} Gradient-Free TI \cite{73fei2023gradient} & Object  & TTF                & SD           & Evolutionary Strategy                                         \\\hline
\rowcolor{c1} PerSAM \cite{21zhang2023personalize}             & Object  & TTF               & SD            & Mask-assisted Generation                                      \\\hline
\rowcolor{c1} DisenBooth \cite{18chen2023disenbooth}           & Object  & TTF               & SD 2.1          & Regularization                                              \\\hline
\rowcolor{c1} NeTI \cite{61alaluf2023neural}                   & Object  & TTF                 & SD 1.4       & Token Embedding Enhancement                                  \\\hline
\rowcolor{c1} Prospect \cite{30zhang2023prospect}              & Object  & TTF                 & SD 1.4          & Token Embedding Enhancement                               \\\hline
\rowcolor{c1} Break-a-Scene \cite{74avrahami2023break}         & Object  & TTF              & SD 2.1            &\makecell[l]{ Attention-based Operation; \\Attention-based Operation;  \\ Mask-assisted Generation   }                              \\\hline
\rowcolor{c1} COMCAT \cite{75xiao2023comcat}                   & Object  & TTF             & SD 1.4        & Attention-based Operation                                       \\\hline
\rowcolor{c1} ViCo \cite{20hao2023vico} & Object  & TTF            & SD           & \makecell[l]{Attention-based Operation; \\Regularization   }                              \\\hline
\rowcolor{c1} OFT \cite{59qiu2024controlling}                  & Object  & TTF              & SD 1.5             & Finetuning Strategy                                       \\\hline
\rowcolor{c1} LyCORIS \cite{76yeh2023navigating}               & Object  & TTF             & SD 1.5          & Attention-based Operation                                     \\\hline
\rowcolor{c1} He et al. \cite{57he2023data}                     & Object  & TTF          & SD, SDXL           & Data Augmentation                                                           \\\hline
\rowcolor{c1} DIFFNAT \cite{78roy2023diffnat}                  & Object  & TTF               & SD            & Regularization                                                \\\hline
\rowcolor{c1} MATTE \cite{79agarwal2023image}                      & \makecell[c]{Object; Style;\\ High-level Semantic}         & TTF          & SD             & Regularization                            \\\hline
\rowcolor{c1} LEGO \cite{45motamed2023lego}                    & Object  & TTF               & SD         & Regularization                                                   \\\hline
\rowcolor{c1} Catversion \cite{80zhao2023catversion}           & Object  & TTF                   & SD 1.5        & Embedding Concatenation                                   \\\hline
\rowcolor{c1} CLiC \cite{81safaee2023clic}                     & Object  & TTF            & SD 1.4            & \makecell[l]{Attention-based Operation; \\Mask-assisted Generation       }                                  \\\hline
\rowcolor{c1} HiFi Tuner \cite{48wang2023hifi}                 & Object  & TTF            & SD 1.4                  & \makecell[l]{Attention-based Operation;\\ Mask-assisted   Generation;  \\ Regularization }                                     \\\hline
\rowcolor{c1} InstructBooth \cite{56chae2023instructbooth}     & Object  & TTF          & SD 1.5                   & Reinforcement Learning                     \\\hline
\rowcolor{c1} DETEX \cite{85cai2023decoupled}                  & Object  & TTF         & SD 1.4                  & Mask-assisted Generation                                  \\\hline
\rowcolor{c1} DreamDistribution \cite{114zhao2023dreamdistribution}     & Object  & TTF     & SD 2.1                               & Token Embedding Enhancement                      \\\hline
\rowcolor{c1} DreamTuner \cite{86hua2023dreamtuner}            & Object  & TTF             & SD        & Attention-based Operation                                           \\\hline
\rowcolor{c1} Xie et al. \cite{51lu2024object}             & Object  & TTF          & SD           & \makecell[l]{Regularization;\\ Mask-assisted Generation }                                   \\\hline
\rowcolor{c1} Infusion \cite{157zeng2024infusion}             & Object  & TTF        & SD 1.5              & Attention-based Operation          \\\hline
\rowcolor{c1} Pair Customization \cite{160jones2024customizing}           & Object  & TTF             & SDXL      & Disentanglement Approach           \\\hline
\rowcolor{c1} DisenDreamer \cite{171chen2024disendreamer}             & Object  & TTF      & SD 2.1     & Disentanglement Approach       \\\hline

\rowcolor{c1} DreamBooth$++$ \cite{219fan2024dreambooth}             & Object  & TTF              & SD         & Attention-based Operation       \\\hline
\rowcolor{c1} CoRe \cite{218wu2024core}             & Object  & TTF             & SD 2.1      & Regularization           \\\hline
\rowcolor{c1} DCI \cite{216jin2025customized}             & Object  & TTF              & SD          & Token Embedding Optimization      \\\hline
 \rowcolor{c1} Re-Imagen \cite{67chen2022re}                    & Object  & PTA             & Imagen        & Retrieval-augmented Paradigm                                                      \\\hline
\rowcolor{c1} Versatile Diffusion \cite{109xu2023versatile}     & Object  & PTA              & SD 1.4           & Architecture Design                                             \\\hline
\rowcolor{c1} Tuning-Encoder \cite{12gal2023encoder}           & Object  & PTA                    & SD      & Regularization                                                     \\\hline
\rowcolor{c1} ELITE \cite{19wei2023elite}                      & Object  & PTA              & SD 1.4              & \makecell[l]{ Attention-based Operation;\\ Mask-assisted Generation    }                                     \\\hline
\rowcolor{c1} UMM-Diffusion \cite{71ma2023unified}             & Object  & PTA                 & SD 1.5        & Reference Feature Injection                                     \\\hline
\rowcolor{c1} SuTI \cite{11chen2024subject}                    & Object  & PTA             & Imagen            & Data Augmentation                                                             \\\hline
\rowcolor{c1} InstantBooth \cite{22shi2023instantbooth}        & Object  & PTA                  & SD 1.4       & Patch Feature Extraction                                        \\\hline
\rowcolor{c1} BLIP-Diffusion \cite{10li2024blip}               & Object  & PTA          & SD 1.5                  &  \makecell[l]{Data Augmentation;\\ Mask-assisted Generation }                         \\\hline
\rowcolor{c1} Domain-Agnostic \cite{17arar2023domain}          & Object  & PTA               & SD           & \makecell[l]{ Regularization; \\Attention-based Operation         }                         \\\hline
 \rowcolor{c1} IP-Adapter \cite{40ye2023ip}                     & Object  & PTA             & SD 1.5              & Reference Feature Injection                                   \\\hline
\rowcolor{c1} Kosmos-G \cite{77pan2023kosmos}                  & Object; Face   & PTA              & SD 1.5           & Multimodal Language Modeling                                    \\\hline
\rowcolor{c1} Kim et al. \cite{47kim2023personalized}    & Object  & PTA             & SD 1.4             & Reference Feature Injection                                    \\\hline
\rowcolor{c1} CAFÉ \cite{84zhou2023customization}              &  Object; Face   & PTA          & SD 2.0               & Multimodal Language Modeling                                    \\\hline
\rowcolor{c1} SAG \cite{99pan2023towards}                      &  Object; Face; Style                & PTA                 & SD 1.5               & Gradient-based Guidance                                  \\\hline
\rowcolor{c1} BootPIG \cite{41purushwalkam2024bootpig}         & Object  & PTA              & SD           & Data Augmentation                                                   \\\hline
\rowcolor{c1} CustomContrast \cite{217chen2024customcontrast}         & Object  & PTA              & SD 1.5            & Contrastive Learning                                               \\\hline
\rowcolor{c1} MoMA \cite{song2024moma}         & Object  & PTA                 & SD 1.5          & Multimodal LLM Adapter                                            \\\hline
\rowcolor{c1} Diptych Prompting \cite{209shin2024large}        & Object; Style  & PTA            & FLUX               &  \makecell[l]{Attention-based Operation;\\ Mask-assisted Generation }                                             \\\hline
 \rowcolor{c2} StyleDrop \cite{5sohn2024styledrop}             &  \makecell[c]{Style;\\ Multiple Subjects}    & TTF             & Muse                  & Data Augmentation                                                     \\\hline
\rowcolor{c2} StyleBoost \cite{53park2023styleboost}           & Style   & TTF           & SD 1.5            & Regularization Dataset                                        \\\hline
\rowcolor{c2} StyleAligned \cite{98hertz2023style}             & Style   & TTF             & SDXL             & Regularization                                                             \\\hline
\rowcolor{c2} GAL   \cite{151zhang2024generative}            &  Style; Object  & TTF            & SD 1.5       & Data Augmentation                                                 \\\hline
\rowcolor{c2} StyleForge \cite{153park2024text}           & Style   & TTF              & SD 1.5            & Parameter-efficient Fine-tuning                                             \\\hline
\rowcolor{c2} FineStyle	\cite{213zhangfinestyle} &Style &TTF&Muse 	&Mask-assisted Generation	\\\hline
\rowcolor{c2} Style-Friendly \cite{211choi2024style} &Style	&TTF& \makecell[c]{FLUX-dev;\\ SD 3.5} 	&SNR Sampler	\\\hline
\rowcolor{c2} StyleAdapter \cite{26wang2023styleadapter}       & Style   & PTA           & SD                      &  \makecell[l]{Data Augmentation;\\ Reference Feature Fxtraction  }                                          \\\hline
\rowcolor{c2} ArtAdapter \cite{42chen2023artadapter}           & Style   & PTA          & SD 1.5             & Data Augmentation                                                 \\\hline
 \rowcolor{c3} ProFusion \cite{55zhou2023enhancing}             & Face    & TTF          & SD 2.0        & Fusion Sampling                                                    \\\hline
\rowcolor{c3}Celeb Basis \cite{60yuan2023inserting}           & Face    & TTF                & SD       & Face Feature Representation                                       \\\hline
 \rowcolor{c3}Banerjee et al. \cite{39banerjee2023identity} & Face    & TTF               & SD 1.4         & Regularization                                               \\\hline
\rowcolor{c3} Magicapture \cite{90hyung2023magicapture}        & \makecell[c]{Face;\\ Multiple Subjects}& TTF          & SD 1.5             &  \makecell[l]{Attention-based Operation;\\ Mask-assisted Generation;\\ Identity Loss}                                                \\\hline
\rowcolor{c3} Concept-centric \cite{35cao2023concept}  & Face    & TTF          & SD 1.5          & Modified Classifier-free Guidance                                \\\hline
\rowcolor{c3} Cross Initialization \cite{66pang2023cross}      & Face    & TTF          & SD 2.1           & Regularization                                                  \\            \hline  
    \end{tabularx}
\end{table*}
\FloatBarrier

\FloatBarrier
\begin{table*}[htb]
    \centering 
    \tiny
    \caption{Paper summary on personalization.} 
    \label{tb:summary2}
\begin{tabularx}{\textwidth}{@{}| X |>{\centering\arraybackslash}p{0.14\textwidth} |>{\centering\arraybackslash}p{0.1\textwidth} |>{\centering\arraybackslash}p{0.13\textwidth} |p{0.28\textwidth}  |@{}}
    \hline
 \textbf{Paper} & \textbf{Scope} & \textbf{Framework} & \textbf{Technique} &\textbf{Backbone}\\ \hline
\rowcolor{c3} OMG \cite{115kong2024omg}              & \makecell[c]{Face;\\ Multiple Subjects}     & TTF          & SDXL             &  \makecell[l]{Attention-based Operation;\\ Mask-assisted Generation;\\ Identity Loss}                                                               \\\hline
\rowcolor{c3} InstantFamily \cite{161kim2024instantfamily}      &  \makecell[c]{Face;\\ Multiple Subjects;\\ Extra Conditions }   & TTF            & SD 1.5             & Mask-assisted Generation                                           \\            \hline  
\rowcolor{c3} PersonaMagic \cite{203li2024personamagic}             & Face    & TTF     & SD 1.4   & Attention-based Operation  \\    \hline
\rowcolor{c3} HyperDreamBooth \cite{58ruiz2023hyperdreambooth} & Face    & TTF; PTA  & SD 1.5 & Pretraining and Fast Fine-tuning                                  \\\hline
\rowcolor{c3} Su et al. \cite{49su2023identity}                 & Face    & PTA       & LDM; StyleGAN     &  \makecell[l]{Identity Loss;\\Multi-task Learning}                 \\\hline
\rowcolor{c3} FastComposer \cite{54xiao2023fastcomposer}       & Face    & PTA            & SD 1.5               &  \makecell[l]{Attention-based Operation;\\ Mask-assisted Generation}                                          \\\hline
\rowcolor{c3} Face0 \cite{88valevski2023face0}                 & Face    & PTA               & SD 1.4          & Fusion Sampling                                                 \\\hline
\rowcolor{c3} DreamIdentity \cite{89chen2023dreamidentity}     & Face    & PTA              & SD 2.1                  & Attention-based Operation                          \\\hline
\rowcolor{c3} Face-Diffuser \cite{52wang2023high}              & Face    & PTA             & SD 1.5               &  \makecell[l]{Attention-based Operation;\\ Mask-assisted Generation}                                         \\\hline
\rowcolor{c3}W-Plus-Adapter \cite{91li2023stylegan}           & Face    & PTA       &SD 1.5; StyleGAN                & Face Feature Representation                            \\\hline
\rowcolor{c3} Portrait Diffusion \cite{32liu2023portrait}      & Face    & PTA              & SD 1.5         & Mask-assisted Generation                                          \\\hline
 \rowcolor{c3} RetriNet \cite{92tang2023retrieving}             & Face    & PTA             & SD            & Mask-assisted Generation                                            \\\hline
\rowcolor{c3} FaceStudio \cite{31yan2023facestudio}            & Face    & PTA                  & SD           & Attention-based Operation                                       \\\hline
\rowcolor{c3} PVA \cite{27xu2024personalized}                  & Face    & PTA                  & LDM       & Mask-assisted Generation                                                         \\\hline
\rowcolor{c3} DemoCaricature \cite{94chen2023democaricature}   &\makecell[c]{ Face; \\Extra Conditions   }              & PTA                  & SD 1.5             & Regularization                                            \\\hline
\rowcolor{c3} PhotoMaker \cite{28li2023photomaker}             & Face    & PTA           & SDXL                  &  \makecell[l]{Data Collection;\\ Reference Feature Extraction}                                                              \\\hline
\rowcolor{c3} Stellar \cite{95achlioptas2023stellar}           & Face    & PTA          & SDXL                 & Evaluation Prompts and Metrics                                                \\\hline
\rowcolor{c3} PortraitBooth \cite{96peng2023portraitbooth}     & Face    & PTA             & SD 1.5                &  \makecell[l]{Attention-based Operation;\\ Mask-assisted Generation}                                        \\\hline
\rowcolor{c3} InstantID \cite{37wang2024instantid}             & \makecell[c]{Face;  \\  Extra Conditions }                & PTA                & SDXL                & Multi-feature Injection                                     \\    \hline
\rowcolor{c3} ID-Aligner \cite{162chen2024id}             & Face              & PTA             & SD 1.5; SDXL            & Feedback Learning           \\    \hline
\rowcolor{c3} MoA \cite{164ostashev2024moa}             & \makecell[c]{Face;\\ Multiple Subjects}         & PTA        & SD 1.5               &  \makecell[l]{Attention-based Operation;\\ Mask-assisted Generation} \\    \hline
\rowcolor{c3} IDAdapter \cite{166cui2024idadapter}             & Face    & PTA    & SD 2.1    & Mixed Facial Features        \\    \hline
\rowcolor{c3} Infinite-ID \cite{169wu2024infinite}             & Face    & PTA      & SDXL  & Reference Feature Injection        \\    \hline
\rowcolor{c3} Face2Diffusion \cite{170shiohara2024face2diffusion}             & Face    & PTA    & SD    & Multi-feature Injection   \\    \hline
\rowcolor{c3} FreeCure \cite{210cai2024foundation}             & Face    & PTA    & SD 1.5; SDXL   & Mask-assisted Generation       \\    \hline
\rowcolor{c3} Omni-ID \cite{205qian2024omni}             & Face    & PTA      & FLUX    & Face representation enhancement     \\    \hline
 \rowcolor{c4} Custom Diffusion \cite{6kumari2023multi}        & Multiple Subjects    & TTF         & SD 1.4             &  \makecell[l]{Parameter-efficient Fine-tuning;\\ Constrained Optimization}                            \\\hline
\rowcolor{c4} Cones \cite{4liu2023cones}                      & Multiple Subjects    & TTF             & SD 1.4               & Concept Neurons Activation                               \\\hline
\rowcolor{c4} SVDiff \cite{8han2023svdiff}                    & Multiple Subjects    & TTF                & SD                  &  \makecell[l]{Data Augmentation;\\ Attention-based Operation }                                         \\\hline
\rowcolor{c4} Perfusion \cite{7tewel2023key}                  & Multiple Subjects    & TTF           & SD 1.5                  & Regularization                                          \\\hline
\rowcolor{c4} Mix-of-Show \cite{46gu2024mix}                   & Multiple Subjects    & TTF       & SD 1.5                & \makecell[l]{Attention-based Operation; \\Mask-assisted Generation}                                          \\\hline
\rowcolor{c4} Cones-2 \cite{62liu2023cones}                   & Multiple Subjects    & TTF          & SD 2.1             &  \makecell[l]{Attention-based Operation;\\ Regularization;\\ Mask-assisted Generation}                                         \\\hline
\rowcolor{c4} PACGen \cite{34li2023generate}                   & Multiple Subjects    & TTF            & SD 1.4                     & Mask-assisted Generation                            \\\hline
\rowcolor{c4} Compositional Inversion \cite{9zhang2023compositional}  & Multiple Subjects    & TTF          & SD             & \makecell[l]{Attention-based Operation;\\ Regularization}                                 \\\hline
\rowcolor{c4} EM-style \cite{44rahman2024visual}               & Multiple Subjects    & TTF          & SD             &  \makecell[l]{Attention-based Operation;\\ Mask-assisted Generation }                                             \\\hline
\rowcolor{c4} MC$^2$ \cite{154jiang2024mc}               & Multiple Subjects    & TTF          & SD 1.5               & Attention-based Operation                                        \\\hline
\rowcolor{c4} MultiBooth \cite{156zhu2024multibooth}               & Multiple Subjects    & TTF      & SD 1.5                & Mask-assisted Generation                                         \\\hline
\rowcolor{c4} Matsuda et al.  \cite{165matsuda2024multi}               & Multiple Subjects    & TTF                  & LDM        & Mask-assisted Generation                                   \\\hline
\rowcolor{c4} MagicTailor \cite{zhou2024magictailor}             & Multiple Subjects          & TTF      & SD 2.1     &  \makecell[l]{Mask-assisted Generation;\\ Attention-based Operation}     \\\hline
\rowcolor{c4} AnyDoor \cite{14chen2023anydoor}                 & Multiple Subjects    & PTA            & SD 2.1                     & Mask-assisted Generation                                \\\hline
\rowcolor{c4} Subject-Diffusion \cite{16ma2023subject}         & Multiple Subjects    & PTA            & SD 2.0                     &  \makecell[l]{Attention-based Operation;\\ Data Augmentation;\\ Mask-assisted Generation}                                  \\\hline
\rowcolor{c4} CustomNet \cite{29yuan2023customnet}             & \makecell[c]{ Multiple Subjects;\\ Extra Conditions}                      & PTA          & SD                & Multi-condition Integration                                  \\\hline
\rowcolor{c4} MIGC \cite{174zhou2024migc}             & Multiple Subjects          & PTA & SD        &  \makecell[l]{Attention-based Operation;\\ Mask-assisted Generation}      \\\hline
\rowcolor{c4} Emu2 \cite{223sun2024generative}             & Multiple Subjects          & PTA    & SDXL      & In-Context Learning
     \\\hline
\rowcolor{c4} SSR-Encoder \cite{221zhang2024ssr}             & Multiple Subjects          & PTA   & SD 1.5      & Attention-based Operation      \\\hline
\rowcolor{c4} $\lambda$-eclipse \cite{222patel2024lambda}             & Multiple Subjects          & PTA  &Kandinsky v2.2        & Contrasstive Learning \\\hline
\rowcolor{c4} MS-Diffusion \cite{220personalizationms}             & Multiple Subjects          & PTA     & SD  & Attention-based Operation         \\\hline
\rowcolor{c6} ReVersion \cite{63huang2023reversion}            & High-level Semantic                   & TTF            & SD            & Regularization                                                   \\\hline
\rowcolor{c6} Inv-ReVersion \cite{159zhang2024inv}            & High-level Semantic                   & TTF           & SD 1.5              & Regularization                                                 \\\hline
\rowcolor{c6} CusConcept \cite{215xu2024cusconcept}            & High-level Semantic                   & TTF     & SD 2.1                    & Regularization                                                 \\\hline
\rowcolor{c6} ADI \cite{43huang2023learning}                   & High-level Semantic                   & TTF          & SD 2.1                  & Mask-assisted Generation                           \\\hline
\rowcolor{c7} PhotoSwap \cite{15gu2024photoswap}               & Extra Conditions                   & TTF        & SD 2.1             & Attention-based Operation                                       \\\hline
\rowcolor{c7} PE-VITON \cite{38zhang2023two}                   & Extra Conditions                   & TTF               & SD          & Shape and Texture Control                                       \\\hline
\rowcolor{c7} Layout-Control \cite{152chen2024training}                   & Extra Conditions                   & TTF        & SD                & Attention-based Operation                                        \\\hline
\rowcolor{c7} SwapAnything \cite{155gu2024swapanything}                   & Extra Conditions                   & TTF           & SD 2.1          & Mask-assisted Generation                                      \\\hline
\rowcolor{c7} PE-VITON \cite{38zhang2023two}                   & Extra Conditions                   & TTF           & SD              & Shape and Texture Control                                       \\\hline
\rowcolor{c7} Viewpoint Control \cite{163kumari2024customizing}                   & Extra Conditions                   & TTF        & SDXL             & 3D Feature Incorporation \\\hline
 \rowcolor{c7} Prompt-Free Diffusion \cite{110xu2023prompt}      & Extra Conditions                   & PTA          & SD 2.0              & Reference Feature Injection                                      \\\hline
\rowcolor{c7} Uni-ControlNet \cite{111zhao2024uni}              & Extra Conditions                   & PTA             & SD             & Multi-feature Injection                                            \\\hline
\rowcolor{c7} ViscoNet \cite{93cheong2023visconet}             & Extra Conditions                   & PTA        & SD                & Multi-feature Injection                                              \\\hline
\rowcolor{c7} Context Diffusion \cite{112najdenkoska2023context}& Extra Conditions                   & PTA       & LDM                   & Multi-feature Injection                                                         \\\hline
\rowcolor{c7} FreeControl \cite{113mo2023freecontrol}           & Extra Conditions                   & PTA          & \makecell[c]{SD 1.5; SD 2.1; \\SDXL}                & Fusion Guidance                                     \\  \hline  
\rowcolor{c7} Li et al. \cite{168li2024tuning}                   & Extra Conditions                   & PTA        & SD           &  \makecell[l]{Attention-based Operation; \\Mask-assisted Generation} \\\hline
    \end{tabularx}
\end{table*}
\FloatBarrier

\FloatBarrier
\begin{table*}[htb]
    \centering 
    \tiny
    \caption{Paper summary on personalization.} 
    \label{tb:summary3}
\begin{tabularx}{\textwidth}{@{}| X |>{\centering\arraybackslash}p{0.14\textwidth} |>{\centering\arraybackslash}p{0.1\textwidth} |>{\centering\arraybackslash}p{0.13\textwidth} |p{0.28\textwidth}  |@{}}
    \hline
 \textbf{Paper} & \textbf{Scope} & \textbf{Framework} & \textbf{Technique} &\textbf{Backbone}\\ \hline

\rowcolor{c8} Tune-A-Video \cite{150wu2023tune}     & Video                & TTF          & SD 1.4         & Parameter-efficient Fine-tuning                                   \\\hline
\rowcolor{c8} Gen-1 \cite{148esser2023structure}            & Video   & TTF              & LDM           & Multi-feature Injection                                                      \\\hline
\rowcolor{c8}Make-A-Protagonist \cite{135zhao2023make} & Video   & TTF            & LDM          & Attention-based Operation                                                       \\\hline
\rowcolor{c8} Animate-A-Story \cite{136he2023animate}  & Video                & TTF      & SD                & Retrieval-augmented Paradigm                                       \\\hline
\rowcolor{c8} MotionDirector \cite{101zhao2023motiondirector}   & Video               & TTF         & \makecell[c]{ZeroScope;\\ ModelScope}                     & Disentanglement Approach                              \\\hline
\rowcolor{c8} LAMP \cite{102wu2023lamp}  & Video                 & TTF        & SDXL; SD 1.4          & First-frame Conditioned Pipeline                             \\\hline

\rowcolor{c8}VMC \cite{140jeong2023vmc}              & Video                & TTF          & Show-1          & Parameter-efficient Fine-tuning                                                \\\hline
\rowcolor{c8} SAVE \cite{103song2023save}             & Video                 & TTF           & VDM          & Attention-based Operation              \\\hline
\rowcolor{c8} Customizing Motion \cite{104materzynska2023customizing} & Video   & TTF          & ZeroScope         & Regularization        \\\hline
\rowcolor{c8} DreamVideo \cite{142wei2023dreamvideo}       & Video   & TTF          & ModelScope           & Disentanglement Approach                                                  \\\hline
\rowcolor{c8} MotionCrafter \cite{106zhang2023motioncrafter}    & Video                & TTF        & VDM             & Disentanglement Approach              \\\hline
\rowcolor{c8} Customize-A-Video \cite{172ren2024customize}      & Video         & TTF       & ModelScope           & Multi-stage Refinement                             \\\hline
\rowcolor{c8} Magic-Me \cite{173ma2024magic}      & Video         & TTF      & \makecell[c]{SD 1.5; \\AnimateDiff}           & Multi-stage Refinement                      \\\hline
\rowcolor{c8} CustomTTT \cite{202bi2024customttt}      & Video         & TTF        & VDM          & Fine-tuning and Alignment
\\\hline
\rowcolor{c8} PersonalVideo \cite{208li2024personalvideo}      & Video; Face         & TTF           & AnimateDiff      & Identity Injection\\\hline
\rowcolor{c8} CustomVideo \cite{145wang2024customvideo}      & \makecell[c]{Video;\\ Multiple Subjects}              & TTF         & ZeroScope                &  \makecell[l]{Data Augmentation;\\ Attention-based Operation}                                                          \\\hline
\rowcolor{c8} VideoDreamer \cite{138chen2023videodreamer}     &\makecell[c]{Video;\\ Multiple Subjects}              & TTF       & SD 2.1                     & Data Augmentation                                        \\\hline
\rowcolor{c8} VideoAssembler \cite{82zhao2023videoassembler}   & Video   & PTA        & VidRD              & Reference Feature Injection                         \\\hline
\rowcolor{c8} VideoBooth \cite{83jiang2023videobooth}          & Video   & PTA       & VDM               & Reference Feature Injection              \\\hline
\rowcolor{c8} VideoMaker \cite{201wu2024videomaker}           & Video & PTA       & VDM                    & Reference Feature Injection                                     \\\hline
\rowcolor{c8} SUGAR \cite{204zhou2024sugar}           & Video & PTA            & VDM              & Attention-based Operation                                \\\hline
\rowcolor{c8} StyleCrafter \cite{212liu2024stylecrafter}        & Video; Style   & PTA           & SD 1.5         & Reference Feature Injection          \\\hline
\rowcolor{c8} ID-Animator \cite{158he2024id}           & Video; Face & PTA           & AnimateDiff           & Reference Feature Injection                             \\\hline
\rowcolor{c8} ConsisID \cite{207yuan2024identity}           & Video; Face & PTA              & CogVideoX          & frequency decomposition                                        \\\hline
\rowcolor{c8} MotionCharacter \cite{206fang2024motioncharacter}           & Video; Face & PTA             & VDM                      & \makecell[l]{ID-Preserving Optimization;\\ Motion Control Enhancement}                            \\\hline
\rowcolor{c8} DreaMoving \cite{105feng2023dreamoving}           &  \makecell[c]{Video; Face;\\ Extra Conditions} & PTA                 & SD                & Multi-feature Injection            \\\hline

\rowcolor{c9} Magic3D \cite{120lin2023magic3d}          & 3D          & TTF	& Imagen; LDM &Score Distillation Sampling                 \\\hline
\rowcolor{c9} DreamBooth3D \cite{121raj2023dreambooth3d}     & 3D      &    TTF & Imagen      	&Multi-stage Refinement                                             \\\hline
\rowcolor{c9} PAS \cite{122azadi2023text}              & 3D  & PTA & SD &	Text-to-3D-Pose  \\\hline
\rowcolor{c9} StyleAvatar3D \cite{123zhang2023styleavatar3d}    & 3D      &  PTA	 & LDM   &Multi-view Alignment                               \\\hline
\rowcolor{c9} AvatarBooth \cite{124zeng2023avatarbooth}      & 3D      & TTF	    & SD   &Dual Model Fine-tuning                                      \\\hline
\rowcolor{c9} MVDream \cite{126shi2023mvdream}          & 3D      &  TTF	  & SD 2.1     &Score Distillation Sampling                               \\\hline
\rowcolor{c9} Consist3D \cite{127ouyang2023chasing}        & 3D      & TTF	   & SD      &Token Embedding Enhancement                                        \\\hline
\rowcolor{c9}Animate124 \cite{128zhao2023animate124}       & 3D      &TTF	   & SD 1.5 &Multi-stage Refinement                            \\\hline
\rowcolor{c9} Dream-in-4D \cite{130zheng2023unified}      & 3D      &TTF   & SD 2.1  	&Multi-stage Refinement                                 \\\hline
\rowcolor{c9} TextureDreamer \cite{133yeh2024texturedreamer}   & 3D      &TTF	    & LDM  &Texture Extraction                                       \\\hline
\rowcolor{c9} TIP-Editor \cite{134zhuang2024tip}       & 3D      & TTF	 & SD     & \makecell[l]{Multi-stage Refinement;\\ Mask-assisted Generation}                                         \\\hline
\rowcolor{c5} Anti-DreamBooth \cite{13van2023anti}            &  Attack and Defense              & TTF        & SD 2.1                  & Perturbation Learning                                           \\\hline
\rowcolor{c5} Concept Censorship \cite{50zhang2023backdooring} & Attack and Defense              & TTF         & SD 1.4          & Trigger Injection                  \\\hline
\rowcolor{c5} Huang et al. \cite{167huang2024personalization} & Attack and Defense              & TTF        & SD           & backdoor attack             \\\hline
\rowcolor{c10} Continual Diffusion \cite{64smith2023continual}  & Others   & TTF             & SD        & Continual Learning                                                      \\\hline
\rowcolor{c10} SVGCustomization \cite{36zhang2023text}          & Others   & TTF              & SD 1.5      & Fine-tuning and Alignment                                        \\\hline
\rowcolor{c10} StitchDiffusion \cite{25wang2024customizing}     & Others   & TTF            & LDM       & Parameter-efficient Fine-tuning    \\\hline
\rowcolor{c10} MC-TI \cite{214wang2024multi}     & Others   & TTF         & SD 1.5          & Regularization    \\\hline
    \end{tabularx}
\end{table*}
\FloatBarrier

\backmatter





\section*{Acknowledgments}
This work was supported in part by Chinese National Natural Science Foundation Projects U23B2054, 62276254, 62372314, Beijing Natural Science Foundation L221013, InnoHK program, and Hong Kong Research Grants Council through Research Impact Fund (project no. R1015-23).

\section*{Declarations of Conflict of Interest}

The authors declared that they have no conflicts of interest to this work.

\noindent

\bibliographystyle{sn-mathphys}
\bibliography{sn-bibliography}
\begin{figure}[h]%
\centering
\includegraphics[width=0.4\linewidth]{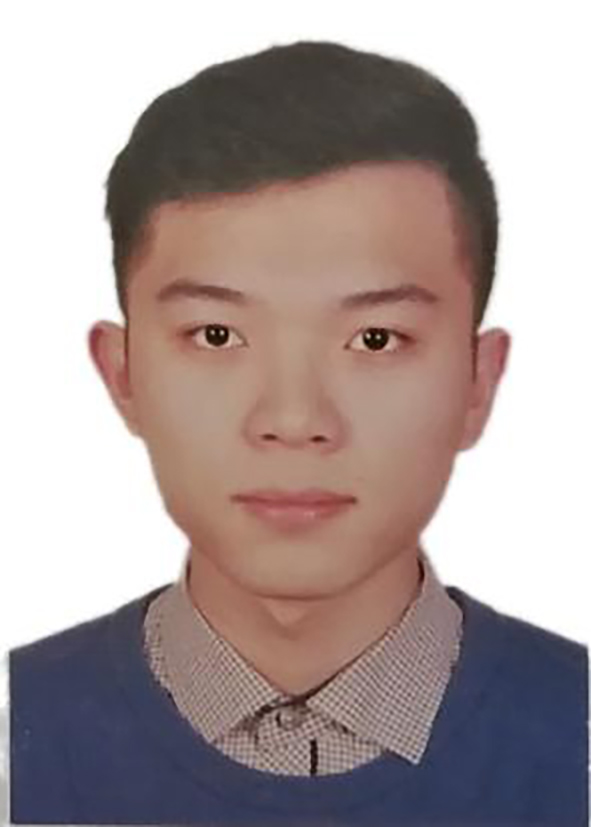}
\end{figure}

\noindent{\bf Xulu Zhang} received the B.E., and M.S. degrees from Sichuan University in 2019 and 2022. He is currently working towards the Ph.D degree from the Department of Computing at the Hong Kong Polytechnic University. His research interests include image generation and active learning.

E-mail: compxulu.zhang@connect.polyu.hk

ORCID iD: 0000-0003-2473-460X


\begin{figure}[h]%
\centering
\includegraphics[width=0.4\linewidth]{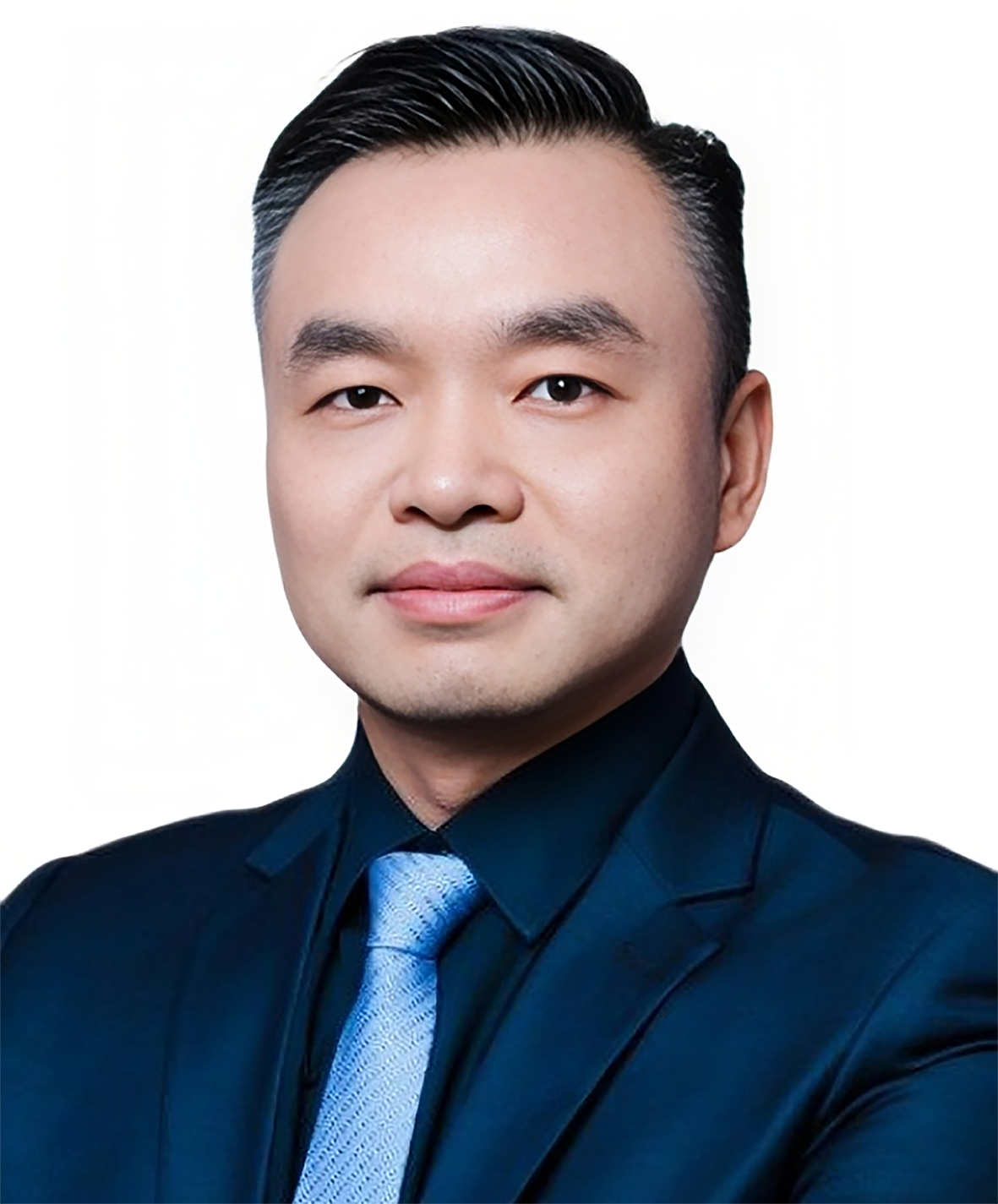}
\end{figure}

\noindent{\bf Xiaoyong Wei}\quad received the Ph.D. degree in computer science from City University of Hong Kong, China in 2009, and has worked as a postdoctoral fellow in the University of California, Berkeley, USA from December, 2013 to December, 2015. He has been a professor and the head of Department of Computer Science, Sichuan University, China since 2010. He is an adjunct professor of Peng Cheng Laboratory, China, and a visiting professor of Department of Computing, Hong Kong Polytechnic University. He is a senior member of IEEE, and has served as an associate editor of Interdisciplinary Sciences: Computational Life Sciences since 2020, the program Chair of ICMR 2019, ICIMCS 2012, and the technical committee member of over 20 conferences such as ICCV, CVPR, SIGKDD, ACM MM, ICME, and ICIP.

His research interests include multimedia computing, health computing, machine learning and large-scale data mining.

E-mail: x1wei@polyu.edu.hk (Corresponding Author)

ORCID iD: 0000-0002-5706-5177


\begin{figure}[h]%
\centering
\includegraphics[width=0.4\linewidth]{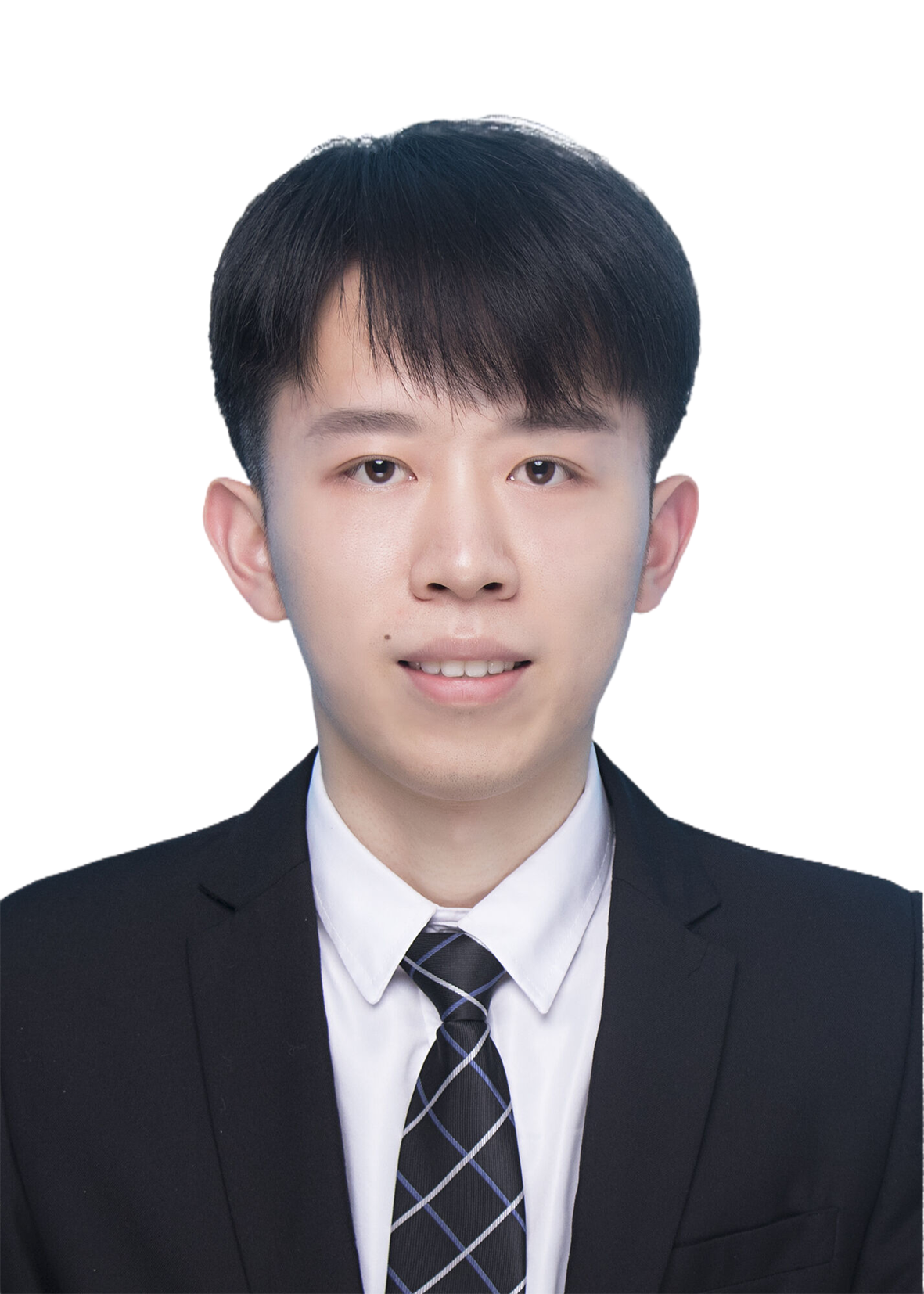}
\end{figure}

\noindent{\bf Wentao Hu}\quad received the B.E., and M.S. degrees from Shandong University and Sun Yat-sen University, in 2021 and 2024. He is currently working towards the Ph.D degree from the Department of Computing at the Hong Kong Polytechnic University. His research interests include image generation and 3D reconstruction.

E-mail: wayne-wt.hu@connect.polyu.hk

ORCID iD: 0000-0002-2071-9341

\begin{figure}[H]%
\centering
\includegraphics[width=0.4\linewidth]{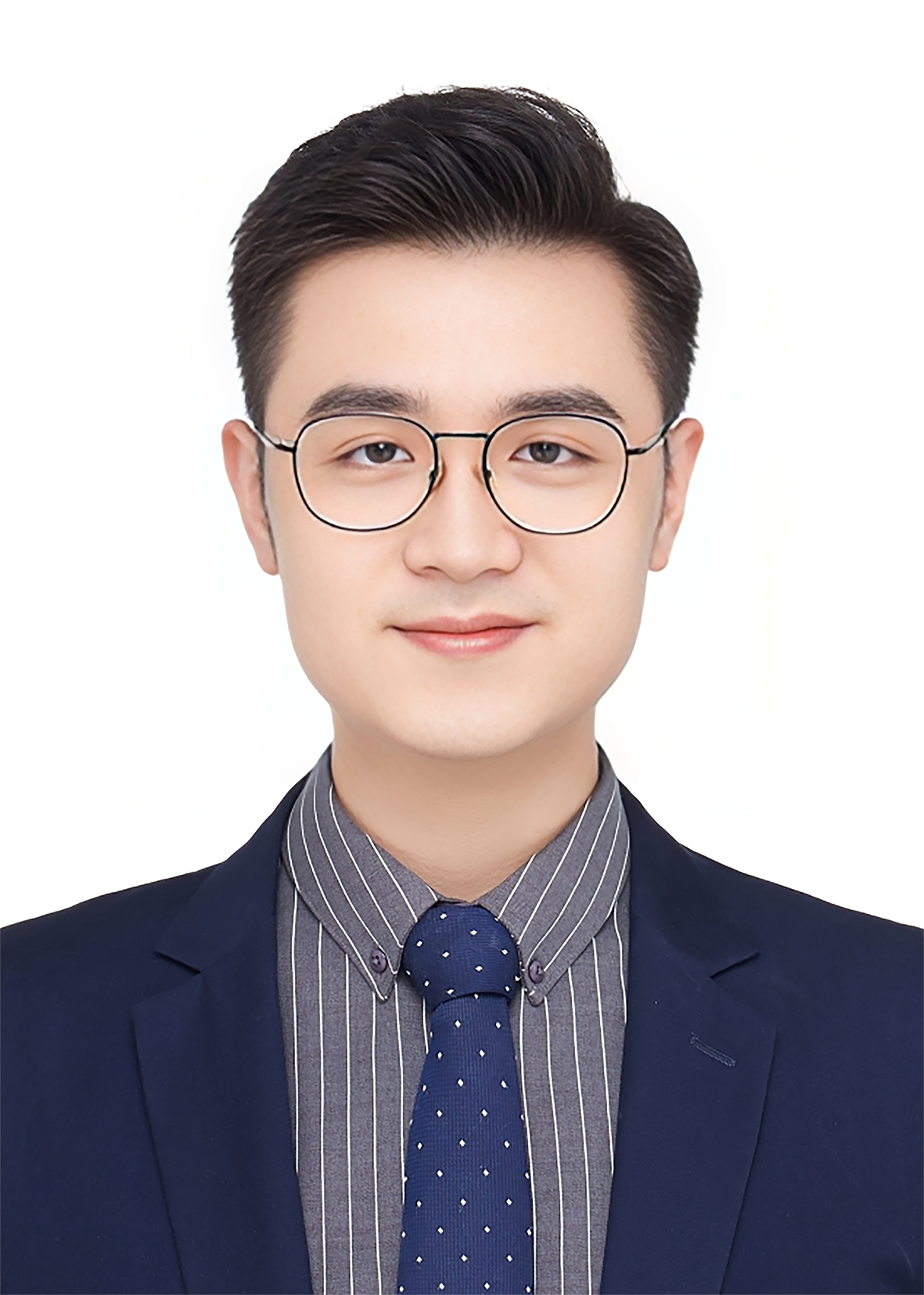}
\end{figure}

\noindent{\bf Jinlin Wu}\quad is an Assistant Research Fellow at the Institute of Automation, Chinese Academy of Sciences (CAS). He received his Ph.D. from the University of Chinese Academy of Sciences in 2022 and his bachelor's degree from the University of Electronic Science and Technology of China in 2017. His research interests include object detection, image recognition, and video understanding. He has served as the principal investigator of a National Natural Science Foundation of China (NSFC) Youth Science Fund project and has participated in several other NSFC-funded projects. Dr. Wu has accumulated a solid research foundation in areas such as video analysis for security and medical video understanding. He has published over 30 high-quality academic papers, with more than 500 citations.

E-mail: jinlin.wu@cair-cas.org.hk 

ORCID iD: 0000-0001-7877-5728 

\begin{figure}[h]%
\centering
\includegraphics[width=0.4\linewidth]{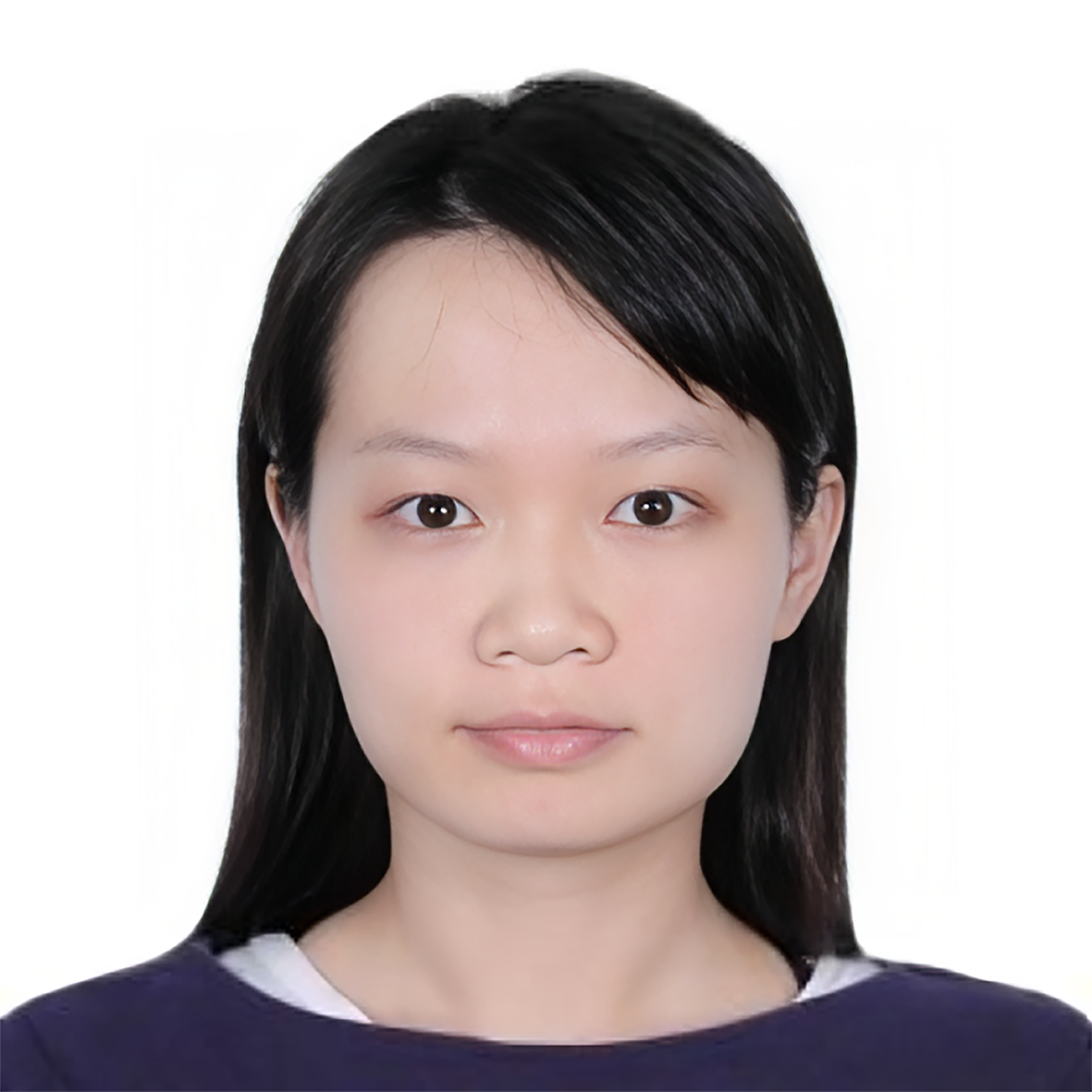}
\end{figure}

\noindent{\bf Jiaxin Wu}\quad is a Postdoctoral Fellow in the Department of Computing at The Hong Kong Polytechnic University. She earned her Ph.D. in Computer Science from the City University of Hong Kong. Her research interests include multimedia retrieval, AI for Science (AI4Science), and natural language processing (NLP).

E-mail: jiaxwu@polyu.edu.hk

ORCID iD: 0000-0003-4074-3442

\begin{figure}[H]%
\centering
\includegraphics[width=0.4\linewidth]{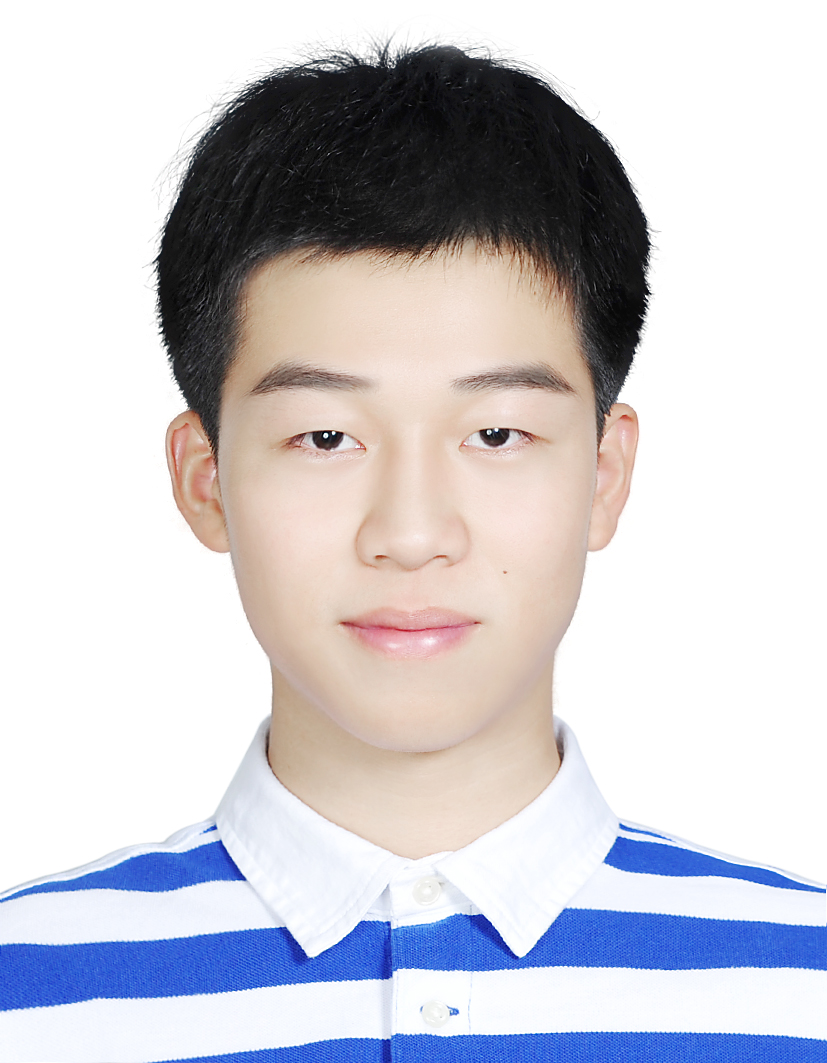}
\end{figure}

\noindent{\bf Wengyu Zhang}\quad is an Undergraduate Student in the Department of Computing at The Hong Kong Polytechnic University. His research interests include natural language processing (NLP), AI for Science (AI4Science), and Graph Learning.

E-mail: wengyu.zhang@connect.polyu.hk

ORCID iD: 0009-0001-2347-4183

\begin{figure}[h]%
\centering
\includegraphics[width=0.4\linewidth]{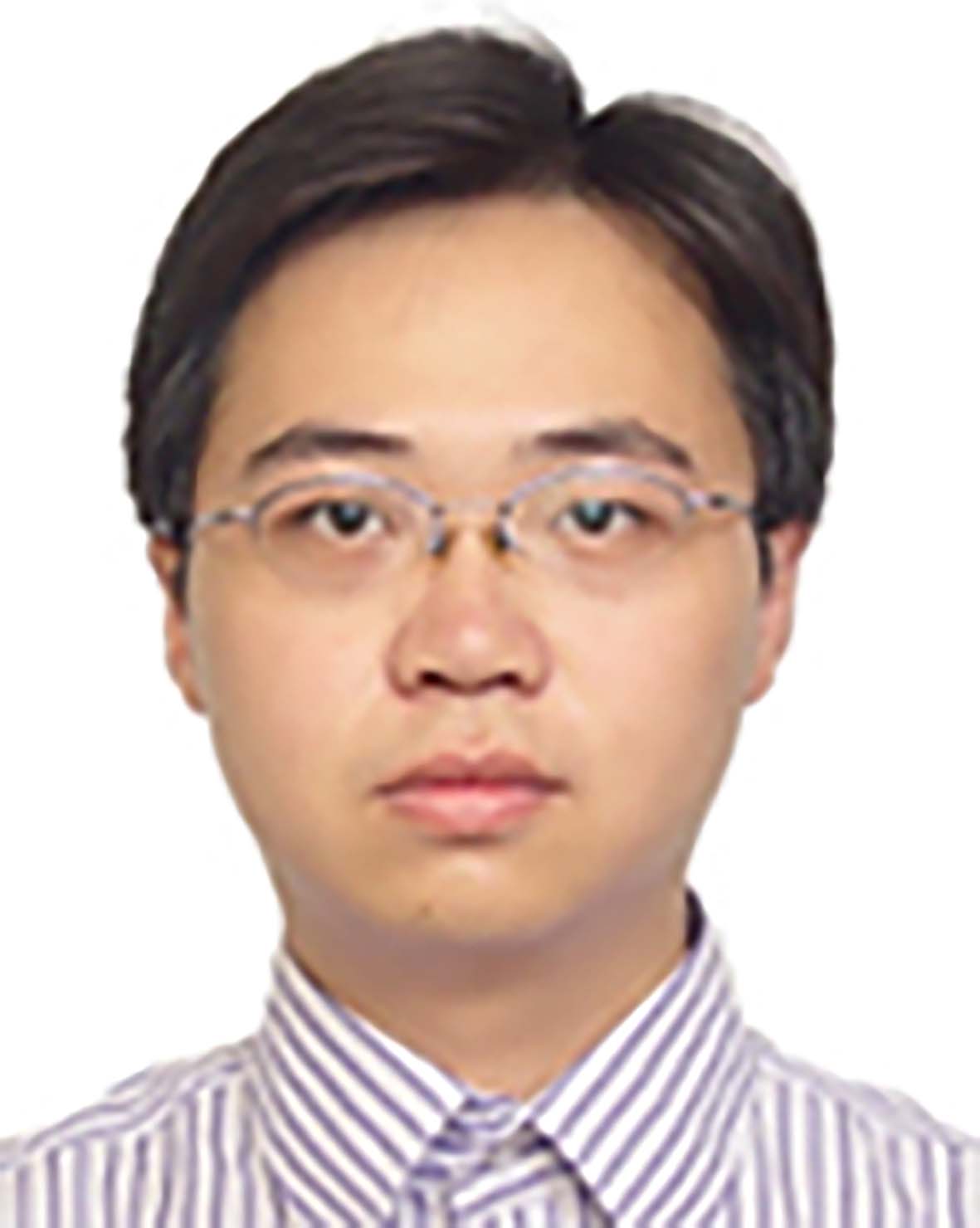}
\end{figure}

\noindent{\bf Zhaoxiang Zhang}\quad received his BSc degree in Department of Electronic Science and Technology from University of Science and Technology of China, and the PhD degree in Pattern Recognition and Intelligent Systems from the Institute of Automation, Chinese Academy of Sciences in 2004 and 2009, respectively. From 2009 to 2015, he worked as a Lecturer, Associate Professor, and later the deputy director of Department of Computer Application Technology at the Beihang University. Since July 2015, Dr. Zhang has joined the Institute of Automation, Chinese Academy of Sciences, where he is currently a professor. His major research interests include pattern recognition, computer vision and machine learning. Recently, he specifically focuses on deep learning models, biologically-inspired visual computing and human-like learning, and their applications on human analysis and scene understanding. He has published more than 200 papers in international journals and conferences, including reputable international journals such as IEEE TPAMI, IJCV, JMLR, IEEE TIP, IEEE TCSVT, and top-level international conferences like CVPR, ICCV, NIPS, ECCV, ICLR, AAAI, IJCAI and ACM MM. He has won the best paper awards in several conferences and championships in international competitions and his research has won the ‘Technical Innovation Award of the Chinese Association of Artificial Intelligence’. He has served as the PC Chair or area chair of many international conferences like CVPR, ICCV, AAAI, IJCAI, ACM MM, ICPR and BICS. He is serving or has served as associate editor of reputable international journals like IJCV, IEEE T-CSVT, IEEE T-BIOM, Pattern Recognition and NeuroComputing.

E-mail: zhaoxiang.zhang@ia.ac.cn

ORCID iD: 0000-0003-2648-3875

\begin{figure}[h]%
\centering
\includegraphics[width=0.4\linewidth]{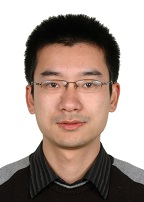}
\end{figure}

\noindent{\bf Zhen Lei}\quad received the B.S. degree in automation from the University of Science and Technology of China, in 2005, and the Ph.D. degree from the Institute of Automation, Chinese Academy of Sciences, in 2010, where he is currently a professor. He is IEEE Fellow, IAPR Fellow and AAIA Fellow. He has published over 200 papers in international journals and conferences with 33000+ citations in Google Scholar and h-index 86. He was the program co-chair of IJCB2023, was competition co-chair of IJCB2022 and has served as area chairs for several conferences and is associate editor for IEEE Trans. on Information Forensics and Security, IEEE Transactions on Biometrics, Behavior, and Identity Science, Pattern Recognition, Neurocomputing and IET Computer Vision journals. His research interests are in computer vision, pattern recognition, image processing, and face recognition in particular. He is the winner of 2019 IAPR Young Biometrics Investigator Award. 

E-mail: zhen.lei@ia.ac.cn (Corresponding Author)

ORCID iD: 0000-0002-0791-189X

\begin{figure}[h]%
\centering
\includegraphics[width=0.4\linewidth]{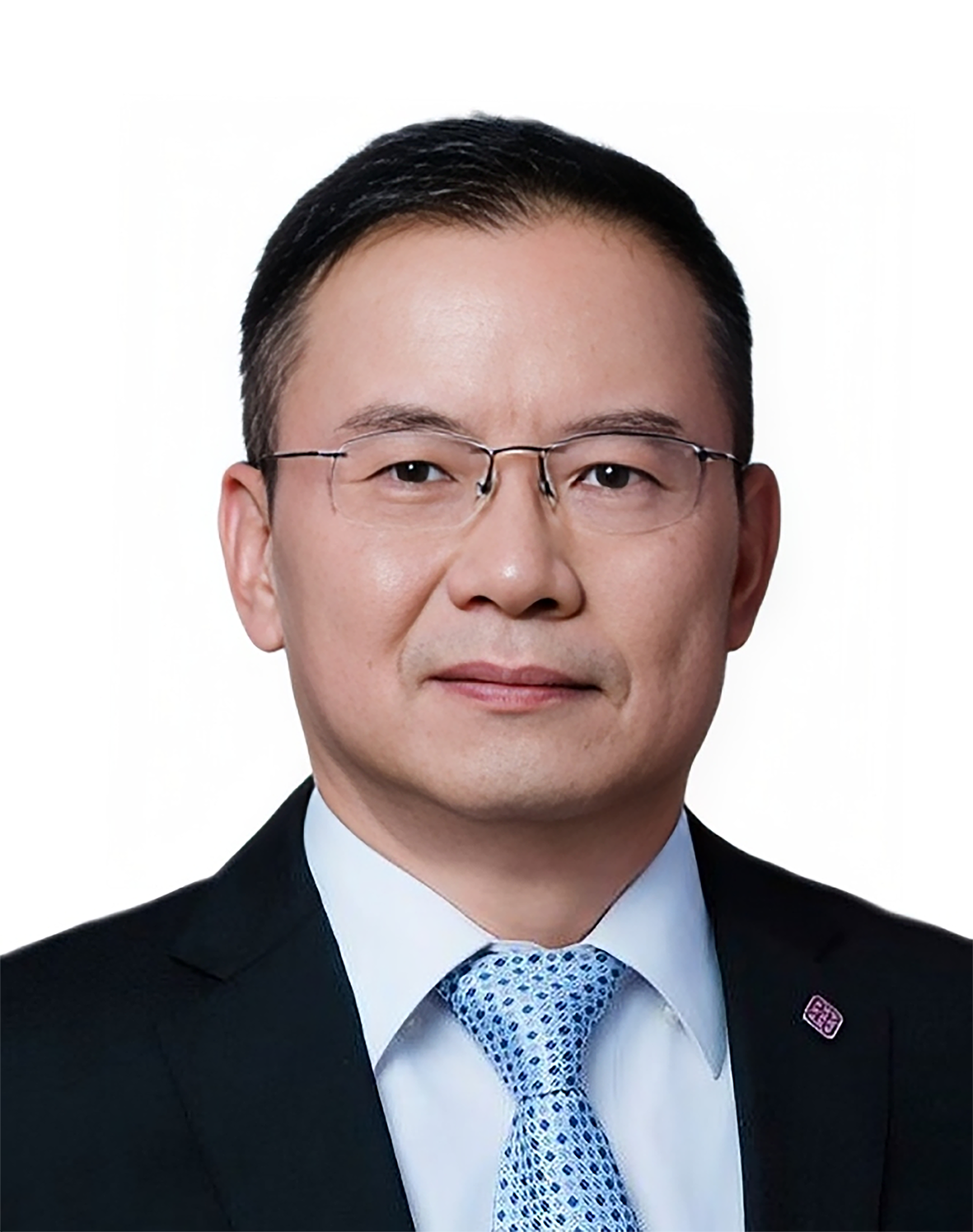}
\end{figure}

\noindent{\bf Qing Li}\quad is currently a Chair Professor (Data Science) and the Head of the Department of Computing, the Hong Kong Polytechnic University. Formerly, he was the founding Director of the Multimedia software Engineering Research Centre (MERC), and a Professor at City University of Hong Kong where he worked in the Department of Computer Science from 1998 to 2018. Prior to these, he has also taught at the Hong Kong University of Science and Technology and the Australian National University (Canberra, Australia). Prof. Li served as a consultant to Microsoft Research Asia (Beijing, China), Motorola Global Computing and Telecommunications Division (Tianjin Regional Operations Center), and the Division of Information Technology, Commonwealth Scientific and Industrial Research Organization (CSIRO) in Australia. He has been an Adjunct Professor of the University of Science and Technology of China (USTC) and the Wuhan University, and a Guest Professor of the Hunan University (Changsha, China) where he got his BEng. degree from the Department of Computer Science in 1982. He is also a Guest Professor (Software Technology) of the Zhejiang University (Hangzhou, China) -- the leading university of the Zhejiang province where he was born.

Prof. Li has been actively involved in the research community by serving as an associate editor and reviewer for technical journals, and as an organizer/co-organizer of numerous international conferences. Some recent conferences in which he is playing or has played major roles include APWeb-WAIM'18, ICDM 2018, WISE2017, ICDSC2016, DASFAA2015, U-Media2014, ER2013, RecSys2013, NDBC2012, ICMR2012, CoopIS2011, WAIM2010, DASFAA2010, APWeb-WAIM'09, ER'08, WISE'07, ICWL'06, HSI'05, WAIM'04, IDEAS'03,VLDB'02, PAKDD'01, IFIP 2.6 Working Conference on Database Semantics (DS-9), IDS'00, and WISE'00. In addition, he served as a programme committee member for over fifty international conferences (including VLDB, ICDE, WWW, DASFAA, ER, CIKM, CAiSE, CoopIS, and FODO). He is currently a Fellow of IEEE and IET/IEE, a member of ACM-SIGMOD and IEEE Technical Committee on Data Engineering. He is the chairperson of the Hong Kong Web Society, and also served/is serving as an executive committee (EXCO) member of IEEE-Hong Kong Computer Chapter and ACM Hong Kong Chapter. In addition, he serves as a councilor of the Database Society of Chinese Computer Federation (CCF), a member of the Big Data Expert Committee of CCF, and is a Steering Committee member of DASFAA, ER, ICWL, UMEDIA, and WISE Society.

E-mail: qing-prof.li@polyu.edu.hk

ORCID iD: 0000-0003-3370-471X

\end{document}